\documentclass{article}

\usepackage{microtype}
\usepackage{graphicx}
\usepackage{subfigure}
\usepackage{booktabs} %

\usepackage{hyperref}

\usepackage[accepted]{icml2025}

\usepackage{amsmath}
\usepackage{amssymb}
\usepackage{mathtools}
\usepackage{amsthm}

\usepackage[capitalize,noabbrev]{cleveref}

\theoremstyle{plain}
\newtheorem{theorem}{Theorem}[section]
\newtheorem{proposition}[theorem]{Proposition}

\newtheorem{corollary}[theorem]{Corollary}
\theoremstyle{definition}

\theoremstyle{remark}

\usepackage[textsize=tiny]{todonotes}

\usepackage{dsfont}

\usepackage{multirow}
\usepackage{colortbl} %
\usepackage{xcolor}
\usepackage{tikz}

\icmltitlerunning{Adaptive Localization of Knowledge Negation for Continual LLM Unlearning}

\newcommand{\method}{\texttt{ALKN}}

\begin{document}
	
\twocolumn[
\icmltitle{Adaptive Localization of Knowledge Negation for Continual LLM Unlearning}

\icmlsetsymbol{equal}{\dag}

\begin{icmlauthorlist}
	\icmlauthor{Abudukelimu Wuerkaixi}{tsinghua}
	\icmlauthor{Qizhou Wang}{hpbu}
        \icmlauthor{Sen Cui}{tsinghua}
	\icmlauthor{Wutong Xu}{tsinghua}
	\icmlauthor{Bo Han}{hpbu,riken}
	\icmlauthor{Gang Niu}{riken}
	\icmlauthor{Masashi Sugiyama}{riken,tokyoU}
        \icmlauthor{Changshui Zhang}{tsinghua}
\end{icmlauthorlist}

\icmlaffiliation{tsinghua}{Institute for Artificial Intelligence, Tsinghua University (THUAI); Beijing National Research Center for Information Science and Technology (BNRist); Department of Automation, Tsinghua University, Beijing, P.R.China}
\icmlaffiliation{hpbu}{TMLR Group, Department of Computer Science, Hong Kong Baptist University}
\icmlaffiliation{riken}{RIKEN}
\icmlaffiliation{tokyoU}{The University of Tokyo}

\icmlcorrespondingauthor{Sen Cui}{cuis@mail.tsinghua.edu.cn}
\icmlcorrespondingauthor{Changshui Zhang}{zcs@mail.tsinghua.edu.cn}

\icmlkeywords{Machine Learning, ICML}

\vskip 0.3in
]

\printAffiliationsAndNotice{}  %

\begin{abstract}
With the growing deployment of large language models (LLMs) across diverse domains, concerns regarding their safety have grown substantially.
LLM unlearning has emerged as a pivotal approach to removing harmful or unlawful content while maintaining utility.
Despite increasing interest, the challenges of continual unlearning, which is common in real-world scenarios, remain underexplored. 
Successive unlearning tasks often lead to intensified utility degradation.
To effectively unlearn targeted knowledge while preserving LLM utility, it is essential to minimize changes in model parameters by selectively updating those linked to the target knowledge, thereby ensuring other knowledge remains unaffected.
Building on the task vector framework, we propose a new method named \method  (Adaptive Localization of Knowledge Negation), which uses dynamic masking to sparsify training gradients and adaptively adjusts unlearning intensity based on inter-task relationships.
Comprehensive experiments across three well-established LLM unlearning datasets demonstrate that our approach consistently outperforms baseline methods in both unlearning effectiveness and utility retention under continual unlearning settings.\footnote{Code is available at: \url{https://github.com/zaocan666/ALKN}}
\end{abstract}

\section{Introduction}
Large Language Models (LLMs) exhibit exceptional capabilities in processing general knowledge, positioning themselves as a promising framework on the path towards artificial general intelligence~\cite{DBLP:journals/corr/abs-2308-12950,wang2024clips,DBLP:conf/nips/BrownMRSKDNSSAA20,DBLP:journals/corr/abs-2303-18223}.
However, reliance on extensive web data introduces risks of memorizing sensitive information, including private data, copyrighted material, and harmful content, hindering lawful deployments of LLMs~\cite{DBLP:journals/corr/abs-2406-08607, wang2025towards}.
These factors continuously spur growing interest in research on LLM unlearning, which aims to remove specified sensitive information from LLMs while retaining their general knowledge and capabilities, i.e., preserving model utility.
It focuses on ensuring that the model cannot reproduce or accurately respond to target data associated with harmful outcomes or legal issues~\cite{DBLP:journals/corr/abs-2402-08787}.

\begin{figure}[t!]
	\vspace{-.1cm}
	\setlength{\abovecaptionskip}{-.1cm}
	\setlength{\belowcaptionskip}{-.3cm}
	\centering{
		\includegraphics[width=\columnwidth]{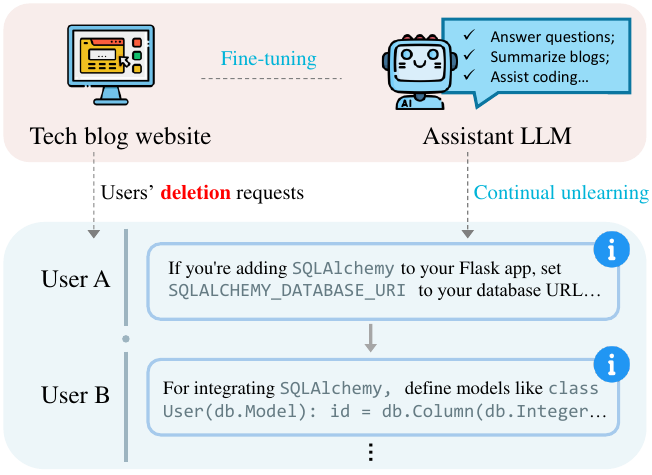}}
        \vspace{-.6cm}
	\caption{An illustrative example of the LLM continual unlearning scenario. An LLM is integrated into a technical blog website, where some users occasionally close their accounts and request the deletion of their blog contents.
    }%
	\label{fig:problem}
	\vspace{-.5cm}
\end{figure}

Although LLM unlearning has received much interest, an important problem of LLM continual unlearning remains underexplored.
In real-world applications, unlearning requests from users or regulators often arrive continuously, requiring models to unlearn multiple pieces of data over time.
For example, we consider a website equipped with a dedicated LLM that has been fine-tuned using user data from the platform, as shown in Figure~\ref{fig:problem}. Periodically, users may request to delete their accounts and have their personal data removed, or regulators may report sensitive information in the LLM that needs to be removed.
In these cases, the LLM needs to continuously perform unlearning tasks to ensure compliance with the requests from users and regulators.

Maintaining model utility represents a critical challenge in LLM unlearning, which is particularly pronounced in the continual unlearning scenario~\cite{DBLP:journals/corr/abs-2407-06460}.
Representative baseline approaches to LLM unlearning use gradient ascent optimization (GA) and its variants~\cite{DBLP:conf/icml/00030YYXTS20, DBLP:journals/corr/abs-2310-10683}, inherently affecting the overall utility of the model due to the extreme modification of model parameters.
The issue of utility deterioration is further magnified in the context of continual unlearning, which can be attributed to two factors.
(I) 
Each unlearning task contributes to a decline in the model utility of the LLM.
And such decline accumulates progressively with an increasing number of unlearning tasks.
We refer to this phenomenon as \textit{accumulative decline}.
(II)
After completing preceding unlearning tasks, the memory retention of the LLM regarding data from subsequent tasks may be disrupted, making them ``partially forgotten" even before they are explicitly unlearned.
Employing a fixed unlearning intensity without adjustment for these partially forgotten tasks can easily result in over-unlearning, further exacerbating the decline in model utility, which we refer to as \textit{cascading degradation}.
While accumulative decline stems from the progressive loss of utility over multiple unlearning tasks, cascading degradation arises from the unintended over-unlearning of partially forgotten data, amplifying the utility loss in subsequent tasks.
These two issues combined exacerbate the utility degradation problem, making continual unlearning in LLMs a significantly more challenging scenario compared to one-time unlearning.

In this paper, we propose \method~(Adaptive Localization of Knowledge Negation) to mitigate the problem of utility degradation in LLM continual unlearning.
\method~is built on the task vector framework, which first fine-tunes a model to reinforce knowledge from unlearning data and then negates the target knowledge by subtracting the parameter offsets from the original model.~\cite{DBLP:conf/iclr/IlharcoRWSHF23}.
Task vectors are trained with a lower-bounded loss, which inherently reduces the risk of excessive unlearning compared to GA.
To further address the accumulative decline and cascading degradation issues in continual unlearning, we propose three key modules incorporated into the fine-tuning process of task vectors.
(I) Entropic soft label loss strategically adjusts unlearning intensity via adaptive training objectives, reducing cascading degradation.
(II) Dynamic gradient sparsity facilitates  selective fine-tuning of  different parameter sets for distinct tasks, minimizing accumulative decline.
(III) Adaptive parameter modulation enables parameter-specific learning rates to further mitigate utility loss.
In summary, the proposed modules, along with the task vector method, effectively address the catastrophic utility decline commonly observed in LLM continual unlearning by dynamically localizing model parameters and adaptively refining gradients.

We conducted comprehensive experiments to validate the effectiveness of our proposed method. 
To objectively assess the utility of models, we introduced \underline{Tr}aining D\underline{a}ta E\underline{v}aluat\underline{i}on Corpu\underline{s} (\texttt{TRAVIS}) generated using the membership inference attacks (MIA) method~\cite{DBLP:conf/iclr/ShiAXHLB0Z24}, designed to reconstruct knowledge in pretraining data of LLMs across a diverse range of topics.
Testing on diverse training data provides a more comprehensive assessment of the overall utility of LLMs and also highlights precise utility decline relative to their pre-unlearning performance~\cite{DBLP:journals/corr/abs-2410-02879}.
Experiments are performed across multiple models and datasets, demonstrating that \method~effectively retains the general utility of the model while achieving effective unlearning of sequential tasks.
After sufficient unlearning, our method retains over 95\% of its model utility, whereas some baseline methods completely lose their utility.
We summarize our key contributions as follows:
\begin{itemize}
    \item We consider a practical yet challenging LLM continual unlearning setting, conducting theoretical and experimental studies on the unique issues of \textit{accumulative decline} and \textit{cascading degradation} in this context.
    \item We introduce a novel method, \method, which adaptively modulates unlearning gradients to effectively minimize the utility degradation that is particularly severe in LLM continual unlearning scenarios.
    \item To rigorously evaluate unlearning methods, we construct an evaluation corpus, \texttt{TRAVIS}, and perform extensive experiments across multiple benchmark datasets. The experimental results validate the effectiveness and demonstrate the superiority of \method~compared to state-of-the-art baselines.
\end{itemize}

\section{Problem Setup and Preliminary}
In this section, we formulate the setting of continual unlearning, introduce the baseline methods GA and task vectors. We also demonstrate the intrinsic causes and negative impact of the cascading degradation issue.
\label{sec:preliminary}
\subsection{LLM continual unlearning}

\textbf{LLM unlearning}.
This paper explores the concept of unlearning in causal LLMs. For a model $\pi_{\theta}$, where $\theta$ denotes the model parameters, the probability of the $(l+1)$-th token given the first $l$ tokens of an input text $x$ is computed by the model followed by a softmax function, formalized as $\pi_{\theta}(x_{[l+1]}|x_{[:l]})$.
The objective of LLM unlearning is to eliminate the influence of a specified target dataset, $D_u=\{x_u^i\}_{i \in [n_u]}$, along with any related generative capabilities, from the model~\cite{DBLP:journals/corr/abs-2404-02062}. 
Sample $x_u$ may take the form of plain text or question-answer pairs.
Crucially, the unlearning process needs to preserve the knowledge and utility of the model on other data.

\textbf{LLM continual unlearning}. In the context of LLM continual unlearning, the goal extends to managing a sequence of unlearning tasks. Beginning with an initial LLM characterized by pre-trained parameters $\theta^0\in \mathbb{R}^d$, the model is required to sequentially unlearn datasets $\{D_u^t\}_{t=1}^T$, where each $D_u^t=\{x_u^{t,i}\}_{i\in{[n_u^t]}}$ corresponds to the $t$-th unlearning task.
During the training of a specific task, the data of other tasks is not accessible.
LLM continual unlearning faces the issue of severe degradation of the model utility on non-target data during successive unlearning processes.
To safeguard the general utility of the model throughout this iterative process, a retain set $D_r=\{x_r^i\}_{i\in [n_r]}$ can be employed to mitigate unintended degradation of utility.

\subsection{Gradient Ascent and Task Vectors}
\label{sec:preliminary_method}
A straightforward approach to addressing the continual unlearning of LLMs is to directly apply existing LLM unlearning methods for each task, such as leveraging gradient ascent-based techniques or task vector methods.

\textbf{Gradient Ascent}.
GA is one of the most fundamental fine-tuning-based approaches for unlearning and often serves as a baseline or a foundational module in many existing methods~\cite{DBLP:journals/corr/abs-2310-10683}. 
While the model acquires knowledge during training through gradient descent, GA facilitates the unlearning process by optimizing in the opposite direction. This is achieved by minimizing the negative cross-entropy loss of predicting the next token on the unlearning dataset:
\begin{equation}
	\min_{\theta} -\frac{1}{n_u}\sum_{x\in D_u}\sum_{l}-\log{\pi_{\theta}(x_{[l+1]}|x_{[:l]})}.
\end{equation}
For simplicity, the normalization of the loss by token length is omitted. 
Unlike gradient descent loss, loss of GA lacks a lower bound, which hinders its ability to converge.
Training with GA could lead to significant deviations in model parameters from their pre-training values, which severely degrades model utility.
The NPO method is also a gradient ascent-based approach. Compared to GA, the gradient in NPO is scaled by a regularization term that gradually diminishes over the course of training, alleviating the problem of over-unlearning to some extent~\cite{DBLP:journals/corr/abs-2404-05868}.

\textbf{Task Vectors}. Introduced by \citet{DBLP:conf/iclr/IlharcoRWSHF23}, the task vector method manipulates the ability of a pre-trained model to address downstream tasks through fine-tuning and the arithmetic operation of model parameters.
When applied to LLM unlearning, the method enables the selective removal of knowledge related to a specific dataset $D_u$.
Specifically, a pre-trained model $\pi_{\theta}$ is fine-tuned on the unlearning dataset $D_u$, resulting in a model $\pi_{\theta_{\text{ft}}}$ with enhanced knowledge of $D_u$. The difference between the fine-tuned parameters and the original parameters, $\theta_{\text{ft}}-\theta$, constitutes the task vector for $D_u$.
By subtracting this task vector from the original parameters $\theta$, a modified model is obtained that has unlearned the knowledge from $D_u$.
Additionally, a scalar $\lambda$ is introduced to control the magnitude of the task vector:
\begin{equation}
	\theta_\text{unlearn}=\theta-\lambda(\theta_{\text{ft}}-\theta).
	\label{eq:tv_arithmetic}
\end{equation}

We theoretically analyze the convergence of the GA and task vector method in Appendix~\ref{sec:convergence_tv_ga}.
In the case of logistic regression unlearning, GA  maintains a gradient magnitude above a positive constant and fails to converge. In contrast, the task vector method is proven to converge, indicating that it is less prone to over-unlearning and may cause less utility degradation in LLMs.
While the task vector method mitigates utility loss to some extent during a single unlearning task, it remains susceptible to accumulative decline and cascading degradation issues in continual unlearning scenarios.

\begin{figure}[t!]
	\vspace{-.1cm}
	\setlength{\abovecaptionskip}{-.1cm}
	\setlength{\belowcaptionskip}{-.3cm}
	\subfigure[Prediction probability under three conditions and model utility.]{
		\centering
		\includegraphics[width=\columnwidth]{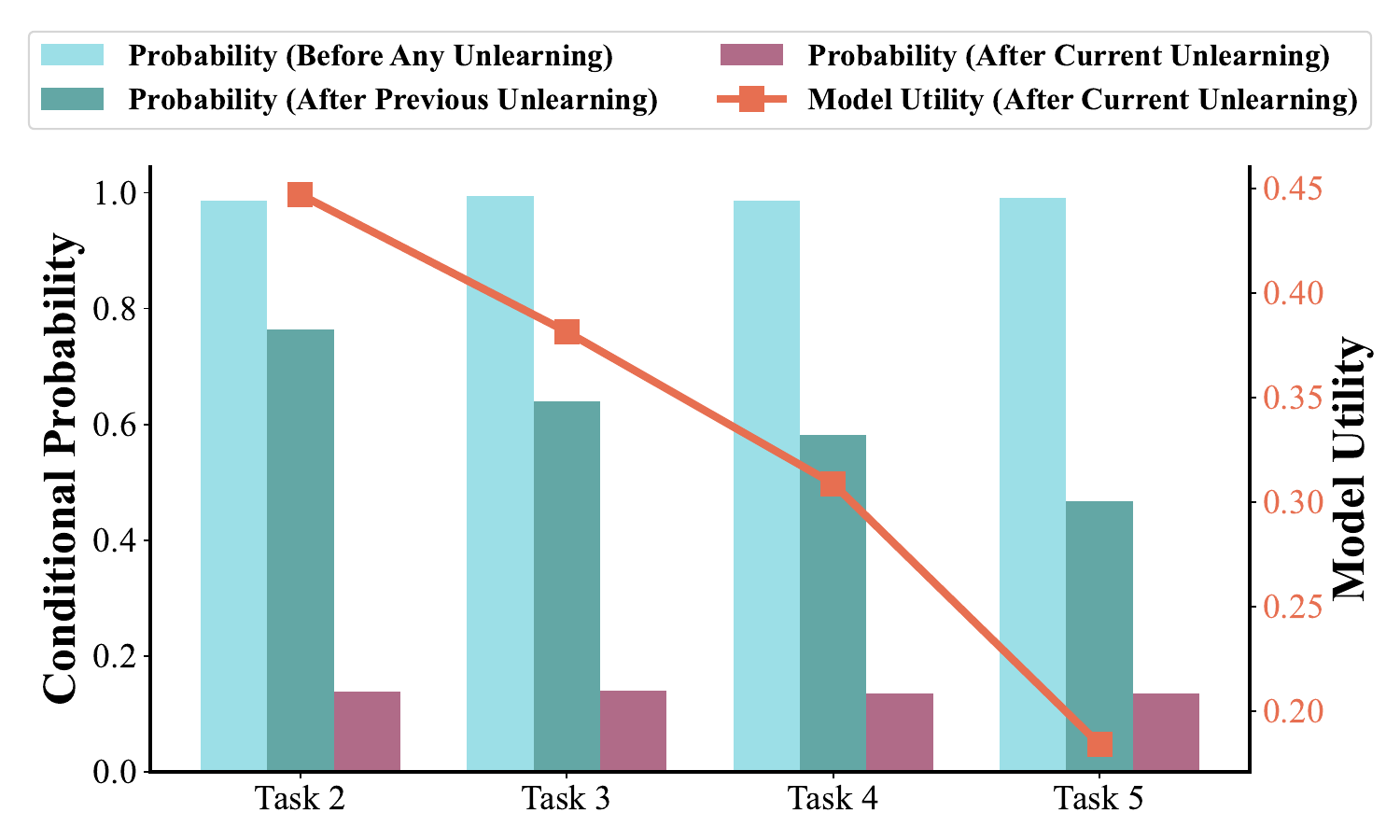}
		\label{fig:cascadingF_prob}
	}
	\subfigure[Rouge-F and utility change under two unlearning scenarios.]{
		\centering
		\includegraphics[width=\columnwidth]{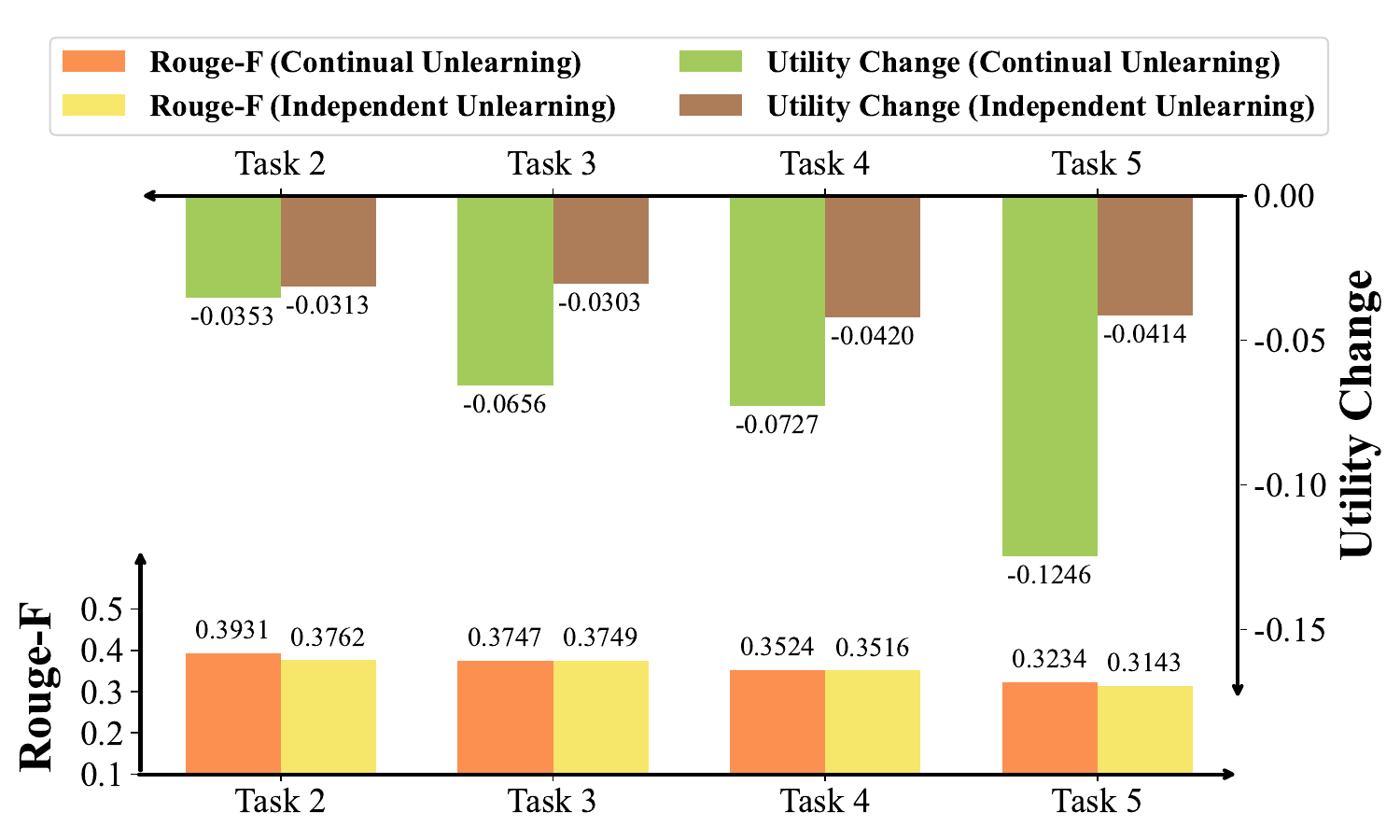}
		\label{fig:cascadingF_utilityChange}
	}
        \vspace{-0.3cm}
	\caption{The comparative experiments on the TOFU dataset using the baseline method to sequentially unlearn five disjoint sub-datasets. (a) The unlearning of preceding tasks results in the partial forgetting of subsequent tasks, as reflected in the reduced conditional probabilities of task data predicted by the model. (b) Under comparable unlearning performance, continual unlearning exhibits significantly greater utility degradation than independent unlearning due to the effects of cascading degradation.
    }
\label{fig:cascadingF}
\vspace{-.3cm}
\end{figure}

\subsection{Cascading Degradation}
\label{sec:cascading_deg}

LLM continual unlearning encounters more complex challenges than one-time unlearning, one of which is cascading degradation, where the interaction between unlearning tasks amplifies the utility degradation.
To elaborate, the utility decline during preceding unlearning tasks can also impair the memory retention of the LLM regarding the data of subsequent tasks.
Besides, in real-world applications, unlearning data is often highly homogeneous, with requests involving similar content.%
As seen in Figure~\ref{fig:problem}, data from different users may share similar structures and content.
This homogeneity intensifies the impact of preceding unlearning tasks on subsequent tasks.
Consequently, the model partially forgets data of subsequent tasks before explicit unlearning, yet these tasks still need to be performed.
Unlearning such partially forgotten data with the same optimization process as preceding tasks can easily lead to over-unlearning, amplifying the utility degradation. We refer to this phenomenon as cascading degradation.

To empirically validate the causes and harm of the cascading degradation, we conducted experiments using a baseline approach, NPO+RT on the TOFU dataset, where the LLM model was made to sequentially unlearn five tasks.
As shown in Figure~\ref{fig:cascadingF_prob}, we examined the model’s prediction probabilities $P(y|x)$ on the data of the $t$-th task before unlearning it (\textcolor[HTML]{218380}{\rule{7pt}{7pt}} Probability After Previous Unlearning).
The results reveal a substantial decrease in these probabilities compared to their initial values (\textcolor[HTML]{73D2DE}{\rule{7pt}{7pt}} Probability Before Any Unlearning), indicating that the model had already partially forgotten the $t$-th task after unlearning the first $t-1$ tasks.
To validate the detrimental effects of cascading degradation, we compared continual unlearning with independent unlearning, where the model independently unlearns the data of each task starting from pre-trained parameters, as shown in Figure~\ref{fig:cascadingF_utilityChange}.
The utility changes in the figure denote utility declines in one unlearning task.
We ensure that the level of unlearning (Rouge-F) achieved by independent
unlearning is similar to that of continual unlearning, enabling a fair comparison of the changes in model utility.
Under equivalent conditions, continual unlearning with cascading degradation leads to a greater and increasingly severe decline in model utility compared to independently unlearning each task.
We also compare the performance of joint unlearning with continual unlearning in Appendix~\ref{sec:more_exp_results} to further validate the harm of cascading degradation.

To theoretically substantiate the above conclusion, we consider a toy example where a logistic model $\pi(y=1|x,\theta)=\sigma(\langle x,\theta\rangle)$ performs unlearning with the GA algorithm on a binary classification problem with inputs being $x$ and labels being $y$.
Suppose there are two unlearning tasks with their dataset being correlated.
Our objective is to verify whether the first unlearning task influences the second task.

\begin{proposition}
Consider two correlated datasets $D^f$ and $D^s$, each composed of $\{(x_i, y_i)\}_{i=1}^n$, satisfying:
$\max_{i\neq j}|\langle x^f_j,x^s_i\rangle|<k\langle x^f_i,x^s_i\rangle$, and $y_i^f=y_i^s$, where $k$ is a positive constant. 
The logistic model starts with an initial parameter $\theta^0$.
We compare two optimization scenarios:
\begin{enumerate}
    \item The model successively unlearns on $D^s$ and $D^f$, yielding intermediate and final parameters: $\theta_s$ and $\theta_{\text{CUL}}$.
    \item Starting from $\theta^0$, the model directly unlearns on $D^s$, yielding parameters $\theta_{\text{IUL}}$.
\end{enumerate}
The parameter changes in the two scenarios during unlearning on $D^s$ satisfy:
\begin{equation}
    ||\bigtriangleup\theta_\text{CUL}||_{X^\top X}>||\bigtriangleup\theta_\text{IUL}||_{X^\top X}+TC\sqrt{n},
\end{equation}
where $C$ is a positive constant depending on the datasets, $T$ is iteration steps, and n is the number of samples.
\label{prop:cul}
\end{proposition}
Please refer to Appendix~\ref{sec:app_proof_cul} for the proof.
Proposition~\ref{prop:cul} demonstrates that during the process of continual unlearning, conducting a preceding unlearning task causes subsequent related tasks to experience larger parameter changes in their own unlearning.
In the context of LLMs, this phenomenon may exacerbate the decline in model utility.

\section{Method}
\label{sec:method}

\begin{figure}[t!]
	\vspace{-.1cm}
	\setlength{\abovecaptionskip}{-.1cm}
	\setlength{\belowcaptionskip}{-.3cm}
	\centering{
		\includegraphics[width=\columnwidth]{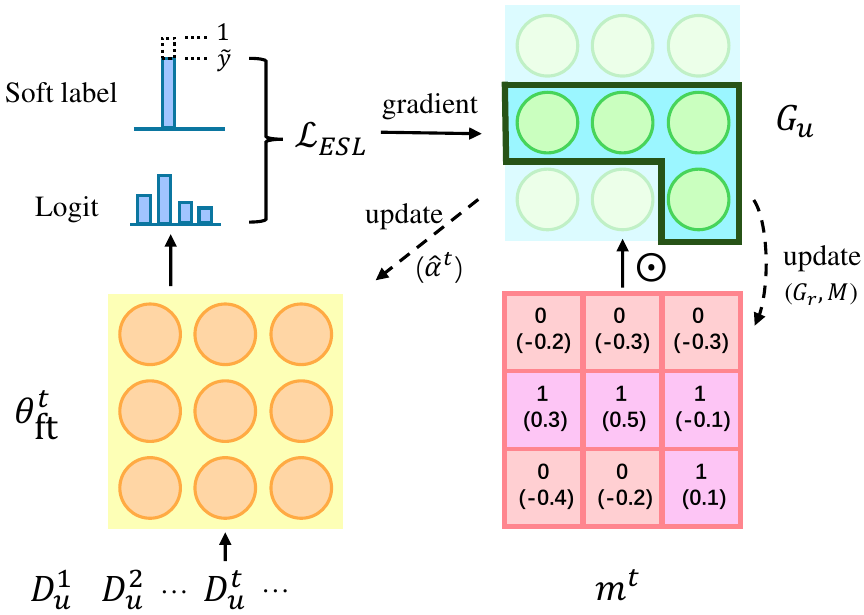}}
        \vspace{-0.4cm}
	\caption{The diagram illustrating the training process of the intermediate model $\pi_{\theta_{\text{ft}}^t}$ for the $t$-th unlearning task. The training loss $\mathcal{L}_{ESL}$ is computed using entropic soft labels (Section~\ref{sec:ESL}). Gradients $G_u$ derived from the loss are masked using a binary mask with an underlying vector $m^t$, dynamically updated based on $G_u, G_r$ and $M$ (Section~\ref{sec:dynamic_gradient}). The masked gradients are then employed to update the model parameters $\theta_{\text{ft}}^t$ with an adaptive learning rate $\hat{\alpha}^t$ (Section~\ref{sec:method_modulation}).}
\label{fig:method}
\vspace{-.3cm}
\end{figure}

In continual unlearning for LLMs, there is severe deterioration in model utility.
To effectively mitigate the utility decline, the key is reducing the overall changes to the model's parameters under the condition of effective unlearning.
Utilizing the task vector method can mitigate utility decline to some extent, but the accumulative decline and cascading degradation issues still exist.
Therefore, we further propose three modules to alleviate these issues.
\textit{Entropic soft label loss} utilizes generated labels to flexibly adjust the unlearning objective. \textit{Dynamic gradient sparsity} applies a learnable mask to the gradients and encourages different tasks to localize distinct model parameters. \textit{Adaptive parameter modulation} employs a lower learning rate for parameters related to already unlearned tasks.
These modules collaboratively address the challenges of \textbf{accumulative decline} and \textbf{cascading degradation} by mitigating task interference and adaptively regulating the intensity of unlearning. The diagram of these modules is shown in Figure~\ref{fig:method}. The overall procedure of training is in Algorithm~\ref{alg:ours}.

\subsection{Fine-tuning with Entropic Soft Labels}
\label{sec:ESL}

To perform the $t$-th unlearning task, we employ knowledge negation from the current model $\pi_{\theta^{t-1}}$ with the task vector method.
This process begins by initializing a model with parameters $\theta^{t-1}$, which is then fine-tuned on the unlearning dataset $D_u^t$. The fine-tuning step yields the intermediate model $\pi_{\theta^t_\text{ft}}$, reinforcing the knowledge of the model specific to the $t$-th dataset~\cite{DBLP:conf/iclr/IlharcoRWSHF23}.
Subsequently, the knowledge associated with the $t$-th task is removed using model weight arithmetic: $\theta^{t}=\theta^{t-1}-\lambda(\theta_{\text{ft}}^{t}-\theta^{t-1})$.

The fine-tuning process for task vector training typically leverages gradient descent optimization with a cross-entropy loss, which quantifies the discrepancy between the next-token prediction labels and the model’s output probability distribution conditioned on preceding tokens:
\begin{equation}
	\mathcal{L}_{CE}(\theta^t_\text{ft}, x)=\sum_{l}-\log{\pi_{\theta^t_\text{ft}}(x_{[l+1]}|x_{[:l]})},
\end{equation}
where $x \in D_u^t$. In the remainder of this paper, $\pi_{\theta}(x_{[l+1]}|x_{[:l]})$ is denoted as $\pi_{\theta}^*$ for simplicity.

However, employing cross-entropy loss in continual unlearning scenarios poses the risk of over-unlearning.
As demonstrated in Section~\ref{sec:cascading_deg}, interaction between unlearning tasks could result in a task being partially forgotten even before its data is explicitly unlearned.
The predicted probability $\pi_{\theta_{\text{ft}}^t}^*$ for the data of the task gets notably low.
In such cases, the cross-entropy loss, which enforces alignment between the output of the model and a one-hot label in the token space, produces gradients as follows:
\begin{equation}
	\nabla_{\theta_{\text{ft}}^t}\mathcal{L}_{CE}\propto -\frac{\nabla_{\theta_{\text{ft}}^t}\pi_{\theta_{\text{ft}}^t}^*}{\pi_{\theta_{\text{ft}}^t}^*}.
\end{equation}
When $\pi_{\theta_{\text{ft}}^t}^*$ is low, the gradient magnitude may increase, causing large parameter updates that negatively impact the utility of the LLM.
This issue is theoretically illustrated in Proposition~\ref{prop:cul}.
Besides, one-hot labels in cross-entropy loss are unnecessarily difficult objectives to optimize towards since the goal of task vector fine-tuning is only to enhance the knowledge of the model regarding the target data.
Training with one-hot labels forces redundant parameter adjustments of the LLM model, leading to significant utility decline.

To mitigate this issue, we propose Entropic Soft Labels (ESL) to preserve model utility during task vector fine-tuning.
We introduce soft labels that increase the entropy of the target distribution in the loss.
The soft label $\tilde{y}$ is derived from the initial predicted probability $\pi_{\theta^{t-1}}^*$ for a training sample and is defined as $\tilde{y}=\sigma\left(s(\pi_{\theta^{t-1}}^*-1)\right)$, where $\sigma$ is the sigmoid function and $s$ is a scaling factor that controls the entropy of labels.
By leveraging these soft labels, the cross-entropy loss is reformulated as:
\begin{equation}
	\mathcal{L}_{ESL}(\theta^t_\text{ft}, x)=\sum_{l}-\tilde{y}_{[l]}\log{\pi_{\theta^t_\text{ft}}(x_{[l+1]}|x_{[:l]})},
\end{equation}
where $\tilde{y}_{[l]}$ represents the soft label for the next token.
$\mathcal{L}_{ESL}$ mitigates the cascading degradation issue by generating adaptive optimization goals for partially forgotten data.

\subsection{Dynamic Gradient Sparsity}
\label{sec:dynamic_gradient}

In continual unlearning scenarios, we aim to fine-tune different sets of model parameters for each task.
This approach helps prevent cumulative parameter divergence from their initial values as the number of tasks increases, thereby reducing the risk of accumulative decline.
Moreover, minimizing changes to model parameters during unlearning is essential for maintaining utility.
We strive to identify parameters that significantly influence the unlearning objective and focus on adjusting these parameters during training.
This would enable effective unlearning while making as few adjustments to the model parameters as possible.

To achieve these goals, we propose to apply learnable masks to model gradients during the fine-tuning of the model $\theta_{\text{ft}}^{t}$ on the unlearning set $D_u^t$, allowing for precise and efficient parameter adjustments.
Inspired by \citet{DBLP:conf/nips/OswaldZKSCZS21}, for the $t$-th task, the optimization of the model parameters is given by:
\begin{equation}
	\theta_{\text{ft}}^{t,k+1}=\theta_{\text{ft}}^{t,k}-\hat{\alpha}^t\odot(\mathds{1}_{m^t\geq\eta}\odot\nabla_{\theta_{\text{ft}}}\mathcal{L}_{ESL}(\theta_{\text{ft}}^{t,k}, D_u^t)),
	\label{eq:theta}
\end{equation}
where $m^t\in \mathbb{R}^d$ represents a dynamic underlying vector, from which a binary mask is derived using the threshold function $\mathds{1}_{m^t\geq\eta}$. This function outputs $1$ for parameters where the corresponding value of $m^t$ exceeds the threshold $\eta$; otherwise it outputs $0$.
The threshold $\eta$ is determined by the percentile of $m^t$ and increases gradually throughout the training process.
Additionally, $\hat{\alpha}^t$ denotes the adaptive learning rate, which is detailed in Section~\ref{sec:method_modulation}.

\textbf{Mask Value Updating}. 
The application of gradient masking is designed to achieve two purposes:
(I) To encourage different tasks to adjust distinct sets of model parameters during unlearning, thereby reducing the overall parameter changes.
(II) To dynamically identify model parameters that are important to the unlearning objective while also being beneficial to utility preservation.
To achieve these goals, the underlying vector $m^t$ is dynamically learned from the data, allowing the mask to adapt to task-specific requirements during training.
For objective (I), the current mask is encouraged to differ from the masks used in previous tasks. Specifically, let $M^{t-1}=\bigcup_{i=1}^{t-1} \mathds{1}_{m^i\geq\eta}$ represent the set of parameters that underwent substantial adjustments in the first $t-1$ tasks.
The mask for the current task is encouraged to minimize its overlap with $M^{t-1}$.
For objective (II), the mask is refined to preserve model utility.
The cross-entropy loss on $D_r$ is used as a proxy to evaluate the retained utility of models. The underlying vector $m^t$ of the mask is optimized to minimize this utility loss.
The sum of these two objectives with a scaling factor $\mu$ is given by:
\begin{equation}
	\min_{m^t}\mathcal{L}_{CE}(\theta^{t-1}-\lambda(\theta_{\text{ft}}^{t,k+1}-\theta^{0}), D_r)+\mu\mathds{1}_{m^t\geq\eta}\cdot M.
\label{eq:mt_obj}
\end{equation}
By substituting into Equation~\ref{eq:theta} and differentiating this objective with respect to $m^t$, we obtain the iterative update rule for $m^t$:
\begin{equation}
\begin{aligned}
	m^t\leftarrow &m^t+\lambda G_u\odot G_r-\mu M,\\
	\text{where}~&~G_u=\nabla_{\theta_{\text{ft}}}\mathcal{L}_{ESL}(\theta_{\text{ft}}^{t,k}, D_u^t), \\
    &~G_r=\nabla_{\theta^{t-1}}\mathcal{L}_{CE}(\theta^{t-1}, D_r).
	\label{eq:mt_it}
\end{aligned}
\end{equation}

The derivation is detailed in Appendix~\ref{sec:derivation_eq_mt_it}.
According to the update rule of $m^t$, the elements will increase where the gradient signs are consistent across both the unlearning objective and the utility objective.
And the corresponding parameters will be selectively activated.
In other words, the proposed method selectively activates model parameters, aligning the gradient directions of the unlearning and utility preservation objectives.
Furthermore, elements that have a greater impact on both objectives—i.e., those with larger gradient magnitudes—lead to faster changes in the corresponding values of $m^t$, resulting in quicker activation or freezing of the model parameters.

The proposed method localizes distinct parameters for different unlearning tasks.
By avoiding cumulative changes to parameters, we alleviate the accumulative decline in utility of the LLM.
And the gradient sparsity helps preserve the overall model utility, further mitigating the issue.

In contrast to methods that rely on static model parameter masks, our proposed approach dynamically adjusts the mask throughout the training process. This dynamic tuning ensures that each model parameter is trained within an appropriate range: parameters with minimal impact on the unlearning objective undergo limited adjustments in the early stages and are subsequently masked, while parameters with significant impact receive more extensive training. This strategy enables balanced training across the model, preventing excessive adjustments to all parameters or overly drastic changes to a small subset. Moreover, our method leverages the underlying vector $m^t$ to aggregate the influence of data on the mask over time, mitigating biases that may arise from extreme values in a single training iteration. By accumulating data-driven insights, the mask adapts more robustly to task-specific requirements, enhancing the stability and effectiveness of the unlearning process.

\subsection{Adaptive Parameter Modulation}
\label{sec:method_modulation}

To further mitigate the decline in model utility, we propose adaptive parameter modulation during fine-tuning $\theta_{\text{ft}}^{t}$.
The gradient masks obtained from our training process represent the relationship between the model parameters and the features of each task.
Consequently, during the training of the current task, we apply a reduced learning rate to the parameters previously activated, thereby preventing the re-unlearning of features already forgotten and alleviating the cascading degradation of utility.
Specifically, let $M^{t-1}=\bigcup_{i=1}^{t-1} \mathds{1}_{m^i\geq\eta}$ denote the model parameters that have undergone significant adjustments in the first $t-1$ tasks. We apply an adaptive learning rate to model parameters while fine-tuning on the $t$-th task as follows:
\begin{equation}
	\hat{\alpha}^t=\alpha_l M^{t-1}+\alpha_h (1-M^{t-1}),
\end{equation}
where $\alpha_l$ denotes a low learning rate for model parameters that are crucial for previous tasks and $\alpha_h$ denotes a relatively high learning rate.

\begin{algorithm}[t!]
   \caption{Continual unlearning method \method}
   \label{alg:ours}
\begin{algorithmic}
   \STATE {\bfseries Input:} Sequential unlearning datasets $\{D_u^t\}_{t=1}^T$, retain set $D_r$, initial model parameters $\theta^0$, hyperparameters $s$ and $\mu$, learning rates $\alpha_l$ and $\alpha_h$, updating steps $T_u$ for each unlearning task.
   \STATE Initialize the overall mask $M=\mathbf{0}$.
   \FOR{$t=1$ {\bfseries to} $T$}
        \STATE Initialize parameters $\theta^{t,0}_\text{ft}=\theta^{t-1}$.
        \STATE Initialize the underlying vector $m^t=\mathbf{0}$.
        \STATE Calculate $G_r=\nabla_{\theta^{t-1}}\mathcal{L}_{CE}(\theta^{t-1}, D_r)$.
        \FOR{$k=1$ {\bfseries to} $T_u$}
            \STATE Update model parameters $\theta_{\text{ft}}^{t,k}$ with Equation~\ref{eq:theta}.
            \STATE Update the underlying vector $m^t$ with Equation~\ref{eq:mt_it}
            \STATE Tune $\eta$ with percentiles of $m^t$ as in Appendix~\ref{sec:tuning_of_eta}.
        \ENDFOR
        \STATE $\theta^{t}=\theta^{t-1}-\lambda(\theta_{\text{ft}}^{t}-\theta^{t-1})$.
        \STATE $M=M\cup\mathds{1}_{m^i\geq\eta}$.
   \ENDFOR
   \STATE {\bfseries Output:} Final parameters $\theta^T$.
\end{algorithmic}
\end{algorithm}

\section{Experiments}
We conduct extensive experiments on various datasets and models to validate the effectiveness of the proposed method.
\subsection{Experimental Settings}
\textbf{Datasets and models}. 
We conduct experiments on three datasets: TOFU, MUSE News, and WHP, each representing scenarios relevant to real-world applications such as privacy preservation and copyright protection.
To simulate a continual unlearning setting, we divid the data designated for unlearning into multiple disjoint subsets based on the structure of the datasets.
These subsets, either from different parts of the same dataset or a mix of datasets, are treated as sequential unlearning tasks.
The experimental details for each dataset are as follows.
1) \textbf{TOFU}~\cite{DBLP:journals/corr/abs-2401-06121} dataset consists of generated fictional author information. We group the information of four authors into one unlearning task, resulting in five or more continual unlearning tasks.
2) \textbf{MUSE News}~\cite{DBLP:journals/corr/abs-2407-06460} comprises BBC news articles and includes four predefined unlearning subsets that serve as continual unlearning tasks.
3) \textbf{WHP} (Who's Harry Potter)~\cite{DBLP:journals/corr/abs-2310-02238,DBLP:journals/corr/abs-2407-06460} involves the unlearning of original text from the Harry Potter series. We treat the original text of three Harry Potter books as three separate unlearning tasks.
Following recent works~\cite{DBLP:journals/corr/abs-2401-06121, DBLP:journals/corr/abs-2410-17509}, we use Llama-2-7B and Phi-1.5 as target models.

\begin{table*}[tbh]
	\centering
	\caption{Unlearning performance (F-Rouge, FQ) and model utility (MU) on the TOFU dataset for the Phi model.}
	\resizebox{\linewidth}{!}{
		\begin{tabular}{c|cc|c|cc|c|cc|c|cc|c|cc|c}
			\toprule
			\multirow{2}{*}{Methods} 
			& \multicolumn{3}{c|}{\textbf{Task 1}} 
			& \multicolumn{3}{c|}{\textbf{Task 2}} 
			& \multicolumn{3}{c|}{\textbf{Task 3}} 
			& \multicolumn{3}{c|}{\textbf{Task 4}} 
			& \multicolumn{3}{c}{\textbf{Task 5}} \\
			& F-Rouge$\downarrow$ & FQ$\uparrow$ & MU$\uparrow$ 
			& F-Rouge$\downarrow$ & FQ$\uparrow$ & MU$\uparrow$ 
			& F-Rouge$\downarrow$ & FQ$\uparrow$ & MU$\uparrow$ 
			& F-Rouge$\downarrow$ & FQ$\uparrow$ & MU$\uparrow$ 
			& F-Rouge$\downarrow$ & FQ$\uparrow$ & MU$\uparrow$ \\
			\hline
			\rowcolor{gray!20}
			Original 
			& 0.9614 & 3.2e-16 & 0.5237
			& 0.9723 & 7.4e-16 & 0.5237
			& 0.9827 & 4.2e-15 & 0.5237
			& 0.9895 & 7.5e-16 & 0.5237
			& 0.9847 & 5.0e-17 & 0.5237 \\
			\hline
			GA+RT
			& 0.5723 & 7.9e-14 & 0.4766
			& 0.4047 & 9.5e-13 & 0.4427
			& \textbf{0.3524} & 1.3e-4 & 0.3494
			& \textbf{0.3095} & 6.5e-5 & 0.2253
			& \textbf{0.1027} & \textbf{9.7e-2} & 0.0812 \\
			GA+KL
			& 0.3963 & 7.8e-4 & 0.3955
			& 0.3896 & 1.5e-5 & 0.4270
			& 0.3923 & 9.1e-3 & 0.3578
			& 0.4311 & 5.2e-5 & 0.2898
			& 0.3108 & 5.9e-4 & 0.2059 \\
			NPO
			& 0.8430 & 1.9e-16 & 0.4836
			& 0.7322 & 4.6e-15 & 0.4675
			& 0.6091 & 8.5e-9 & \textbf{0.4651}
			& 0.4893 & 5.4e-9 & 0.4257
			& 0.4033 & 8.9e-6 & 0.3817 \\
			NPO+RT
			& 0.3803 & 9.4e-3 & 0.4823
			& 0.3931 & 7.6e-5 & 0.4470
			& 0.3747 & 4.8e-4 & 0.3814
			& 0.3524 & 6.8e-3 & 0.3087
			& 0.3234 & 7.9e-3 & 0.1841 \\
			NPO+KL
			& 0.4109 & 6.0e-5 & 0.4729
			& 0.4359 & 1.0e-6 & 0.4465
			& 0.4018 & 6.3e-4 & 0.3957
			& 0.3891 & 9.9e-6 & 0.3350
			& 0.3651 & 8.5e-5 & 0.2789 \\
			DPO+RT
			& 0.4510 & 8.9e-10 & 0.5043
			& 0.4091 & 4.3e-8 & 0.4432
			& 0.3983 & 4.6e-5 & 0.3602
			& 0.3821 & 3.1e-4 & 0.3406
			& 0.3771 & 3.9e-3 & 0.3359 \\
			TV
			& 0.3957 & 4.8e-4 & 0.4825
			& 0.3719 & 5.2e-3 & 0.4578
			& 0.3573 & 8.9e-3 & 0.3866
			& 0.3371 & \textbf{9.8e-3} & 0.3253
			& 0.3028 & 4.2e-2 & 0.2963 \\
			SKU
			& 0.4279 & 3.0e-5 & 0.4927
			& 0.4055 & 5.8e-4 & 0.4100
			& 0.3839 & 6.0e-4 & 0.3330
			& 0.3805 & 6.5e-5 & 0.3024
			& 0.3346 & 7.2e-3 & 0.2216 \\
			EUL
			& 0.7885 & 2.4e-15 & 0.4839
			& 0.5562 & 2.9e-12 & 0.4662
			& 0.5264 & 7.2e-12 & 0.4551
			& 0.5171 & 1.6e-13 & 0.4537
			& 0.5058 & 8.0e-12 & 0.3938 \\
			SO-NPO
			& 0.4525 & 7.3e-8 & 0.4935
			& 0.4316 & 6.2e-6 & 0.4532
			& 0.4168 & 9.9e-7 & 0.3725
			& 0.4153 & 5.0e-5 & 0.3472
			& 0.4002 & 4.1e-4 & 0.3001 \\
			WAGLE
			& 0.7667 & 2.0e-15 & 0.4731
			& 0.5031 & 4.8e-9 & 0.4669
			& 0.4593 & 7.7e-6 & 0.4314
			& 0.3860 & 2.5e-5 & 0.3400
			& 0.3731 & 3.5e-4 & 0.3013 \\
			\hline
			\rowcolor{gray!20}
			Ours~w/o ESL
			& 0.3627 & 3.5e-2 & 0.5016
			& 0.3320 & 3.7e-2 & 0.4413
			& 0.3067 & 7.3e-2 & 0.4170
			& 0.2684 & 1.4e-1 & 0.3807
			& 0.2344 & 5.7e-1 & 0.3351 \\
			\rowcolor{gray!20}
			Ours~w/o DGS
			& 0.3827 & 5.4e-3 & 0.5048
			& 0.3642 & 1.6e-3 & 0.4559
			& 0.3536 & 6.2e-3 & 0.4437
			& 0.3297 & 4.0e-3 & 0.4296
			& 0.3073 & 7.9e-1 & 0.3893 \\
			\rowcolor{gray!20}
			Ours~w/o APM
			& 0.3703 & 1.4e-2 & 0.5128
			& 0.3464 & 8.2e-3 & 0.4784
			& 0.3320 & 4.7e-2 & 0.4524
			& 0.3043 & 7.3e-2 & 0.4367
			& 0.2517 & 2.5e-1 & 0.3768 \\
			\hline
			Ours 
			& \textbf{0.3703} & \textbf{1.4e-2} & \textbf{0.5128}
			& \textbf{0.3671} & \textbf{2.0e-3} & \textbf{0.4853}
			& 0.3681 & \textbf{2.1e-3} & 0.4627
			& 0.3452 & 1.3e-4 & \textbf{0.4578}
			& 0.3314 & 1.1e-2 & \textbf{0.4534} \\
			\bottomrule
		\end{tabular}
	}
	\label{tab:tofu_phi}
\end{table*}

\textbf{Metrics}.
Unlearning algorithms require evaluation from two perspectives: unlearning performance and model utility, both measured using various metrics.
Metrics for unlearning performance include:
1) \textbf{Rouge-L for unlearning (F-Rouge)} measures similarity between the text generated by the model and the correct text~\cite{DBLP:journals/corr/abs-2401-06121, DBLP:journals/corr/abs-2407-06460}.
2) \textbf{Probability (F-Prob)} evaluates the conditional probability $P(\text{answer}|\text{question})$ of model outputs.
3) \textbf{Membership inference attack (MIA)} calculates Min-K\% Prob~\cite{DBLP:conf/iclr/ShiAXHLB0Z24} for unlearning samples to detect whether the post-unlearning model retains the unlearning set.
4) \textbf{Forget Quality (FQ)} quantifies the difference in prediction distributions on the forget set between the post-unlearning model and a retrained model~\cite{DBLP:journals/corr/abs-2401-06121}.
Metrics for model utility include:
1) \textbf{R-Rouge} and 2) \textbf{R-Prob} are also applied to the retain set or holdout set to evaluate the model utility.
3) \textbf{T-Rouge} calculates the Rouge-L recall of completing sentences in the proposed TRAVIS dataset. Testing on inferred  training data provides a more accurate assessment of the impact of unlearning algorithms on the pre-existing knowledge of the model. Additionally, TRAVIS is generated without thematic constraints, resulting in a broad coverage of content, which enables a more comprehensive evaluation of performance.
4) \textbf{Model Utility (MU)} represents the harmonic mean of the above metrics calculated across multiple retain sets in the TOFU dataset~\cite{DBLP:journals/corr/abs-2401-06121}, including the T-Rouge. It provides an aggregate measure of the model utility.

\textbf{Baselines}.
Given the limited research on LLM continual unlearning, the proposed method is primarily compared with prominent existing unlearning approaches. These algorithms are applied sequentially to unlearn the data for each task, thereby achieving continual unlearning. \textbf{Unlearning methods} include GA~\cite{DBLP:journals/corr/abs-2310-10683}, NPO~\cite{DBLP:journals/corr/abs-2404-05868}, and Task Vectors (TV)~\cite{DBLP:conf/iclr/IlharcoRWSHF23}, as introduced earlier. 
We also include DPO~\cite{DBLP:journals/corr/abs-2404-05868}, SO-PO, and SO-NPO~\cite{DBLP:conf/emnlp/JiaZZLRDK024} methods.
Besides, EUL~\cite{DBLP:conf/emnlp/ChenY23} incorporates lightweight unlearning layers into the model and explicitly considers scenarios involving multiple unlearning tasks.
WAGLE~\cite{DBLP:journals/corr/abs-2410-17509} optimizes only the parameters that are crucial to unlearning and retaining objectives during the unlearning process.
When applying these methods in continual unlearning scenario, the learning rate is tuned progressively to mitigate catastrophic collapse as the number of tasks increases.
\textbf{Retaining objectives} are also leveraged with an auxiliary retain dataset to help preserve the model utility in addition to conducting unlearning optimization on the target data.
For example, the Retain Loss (RT) is employed as $-\mathbb{E}_{D_r}\log{\pi_{\theta}(x_{[l+1]}|x_{[:l]})}$, where gradient descent is performed on the retain dataset to optimize the model.
Similarly, KL divergence loss (KL), defined as $\mathbb{E}_{D_r}D(\pi_\theta(\cdot|x)||\pi_{\text{ref}}(\cdot|x))$ , is used to ensure that the logits of the model outputs on the retain dataset remain similar to those of the initial model $\pi_{\text{ref}}$~\cite{DBLP:journals/corr/abs-2404-05868}.

\subsection{Experiments on Privacy-Preservation Unlearning}
In the TOFU dataset, unlearning algorithms are tested with personal information of fictitious authors, where each unlearning task corresponds to different authors. Table~\ref{tab:tofu_phi} presents the unlearning and utility performance of various methods across five continual tasks, showing that utility declines to varying degrees as tasks increase.
A Rouge score of approximately 0.35 on the unlearning set is considered sufficient for effective unlearning~\cite{DBLP:journals/corr/abs-2401-06121}.
NPO+RT achieves reasonable unlearning performance in the first four tasks but suffers over-unlearning in the last, with model utility dropping below 0.2.
Conversely, GA+RT struggles with insufficient unlearning in earlier tasks, followed by over-unlearning later.
NPO, EUL, and WAGLE exhibit severe under-unlearning, with F-Rouge reductions of less than 15\% in the first task compared to the original model.
Although NPO and EUL achieve the highest utility in the third and fourth tasks, this stems from insufficient unlearning.
GA+KL, NPO+KL, and SKU avoid over-unlearning but suffer progressively lower utility.
In contrast, our method effectively balances unlearning and utility, achieving sufficient unlearning while minimizing the decline in model utility.
Notably, it not only achieves the best performance across all metrics on the first two tasks but also surpasses baseline methods from the perspective of multiple tasks.
After completing all five unlearning tasks, our method exhibits less than a 0.06 reduction in utility, effectively mitigating the cumulative utility degradation typically observed in continual unlearning scenario. This highlights its significantly lower impact on utility compared to other approaches.

\textbf{Ablation study.} We present the ablation study results in Table~\ref{tab:tofu_phi}, illustrating the impact of each module on the overall performance of the model.
When the entropic soft label (ESL) module is removed, the model exhibits significant cascading degradation, resulting in severe utility decline during the unlearning of later tasks.
The dynamic gradient sparsity (DGS) module plays a critical role in identifying model parameters that balance utility retention and effective unlearning. Without the DGS module, the overall model utility decreases, and the cumulative utility degradation becomes more pronounced.
Similarly, the absence of the adaptive parameter modulation (APM) module leads to a sustained decline in utility over the course of unlearning.

\begin{figure}[t!]
	\vspace{-.1cm}
	\setlength{\abovecaptionskip}{-.1cm}
	\setlength{\belowcaptionskip}{-.3cm}
	\centering{
		\includegraphics[width=0.75\columnwidth]{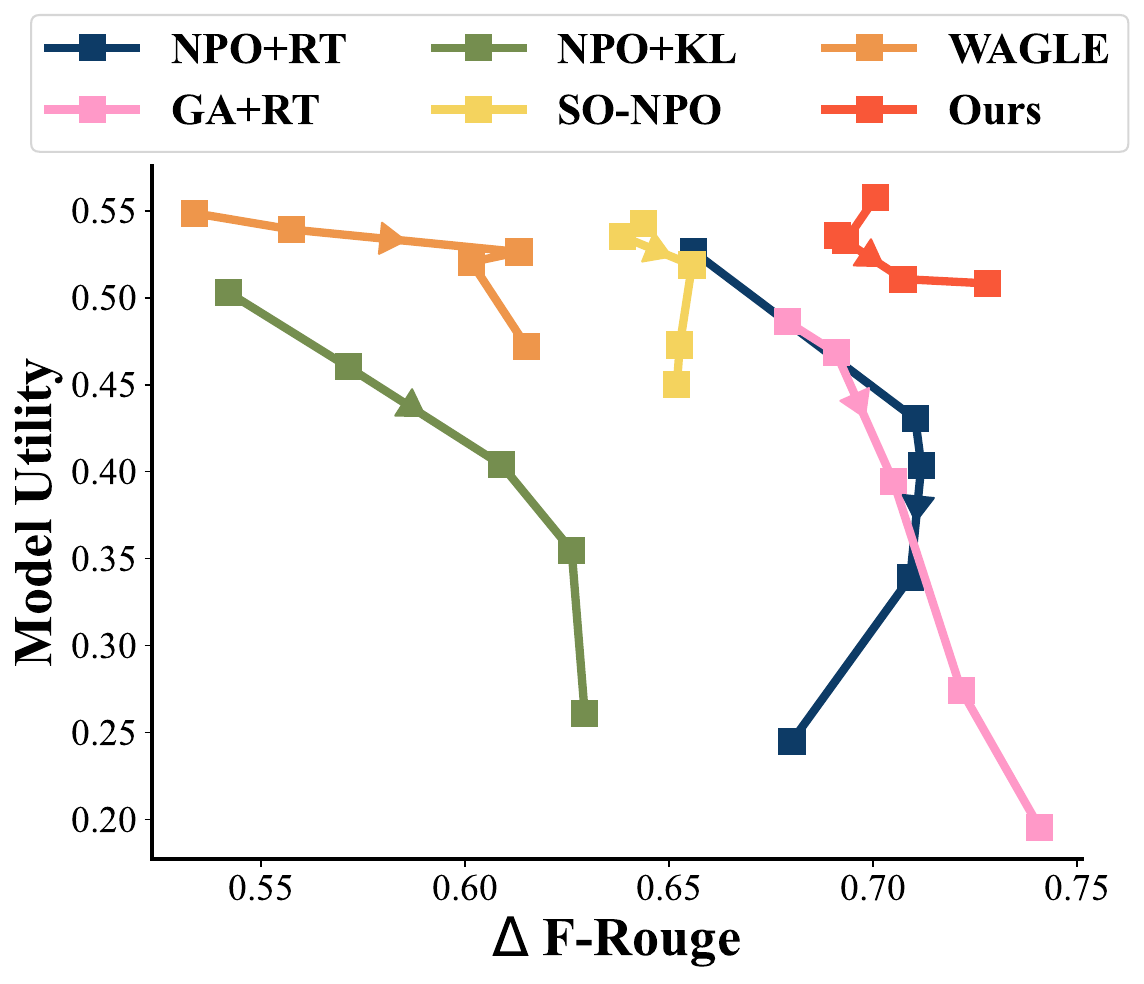}}
	\caption{Model utility (MU) and changes of F-Rouge of five unlearning tasks on TOFU dataset for the Llama2 model.}
	\vspace{-.3cm}
	\label{fig:tofu_llama2_lines}
\end{figure}

Figure~\ref{fig:tofu_llama2_lines} illustrates the unlearning results for the Llama2 model on the TOFU dataset.
The persistent issue of utility degradation is evident, as the data points of the same color representing the performance of successive tasks are generally arranged sequentially from up to down, reflecting the cumulative effect of unlearning tasks on model utility.
For the latter tasks, NPO+RT and GA+RT suffer a drastic decline in model utility. While WAGLE exhibits poor unlearning effectiveness with low changes of F-Rouge.
Notably, the data points corresponding to our method are clustered more tightly across the five tasks, indicating a more stable and controlled utility decline in the continual unlearning scenario.
Furthermore, these points are positioned nearer to the upright corner of the plot compared to other methods, signifying a superior trade-off between effective unlearning and maintaining model utility.
By achieving a high degree of unlearning with minimal impact on utility, our approach demonstrates clear advantages over baseline methods.

\subsection{Experiments on Copyright-Protection Unlearning}
The WHP and MUSE News datasets are designed to evaluate the unlearning of verbatim memory of news articles and Harry Potter books by LLMs. We conducted experiments with the Llama2 model independently on each dataset and further explored the effects of alternating tasks between the two datasets. Please refer to Appendix~\ref{sec:more_exp_results} for more results.

\begin{table}[t!]
	\centering
	\caption{Final model utility metrics and average unlearning performance metrics over three tasks on the WHP dataset.}
	\resizebox{0.97\linewidth}{!}{
		\begin{tabular}{c|cc|cc}
			\toprule
			\multirow{2}{*}{Methods} 
			& \multicolumn{2}{c|}{\textbf{Unlearning}}
			& \multicolumn{2}{c}{\textbf{Utility}} \\
			& F-Rouge$\downarrow$ & MIA$\downarrow$ & R-Rouge$\uparrow$ & T-Rouge$\uparrow$ \\
			\hline
			\rowcolor{gray!20}
			Original
			& 0.9821 & 0.5206 & 0.6940 & 0.7643\\
			\hline
			NPO+RT
			& \textbf{0.1559} & 0.2607 & 0.0 & 0.0\\
			NPO+KL
			& 0.1620 & \textbf{0.2452} & 0.0 & 0.0\\
			GA+RT
			& 0.1855 & 0.2674 & 0.1172 & 0.1975\\			
			SO-NPO
			& 0.3080 & 0.3892 & 0.4109 & 0.4536\\
			WAGLE
			& 0.2946 & 0.3745 & 0.3721 & 0.4594\\
			\hline
			Ours 
			& 0.1790 & 0.2729 & \textbf{0.4586} & \textbf{0.5617}\\
			\bottomrule
		\end{tabular}
	}
	\label{tab:WHP_Trouge}
\end{table}

The results on the WHP dataset are shown in Table~\ref{tab:WHP_Trouge} and Figure~\ref{fig:MUSE_News_WHP_rouge} (b).
Utilizing the \texttt{TRAVIS} dataset provides the evaluation of general utility wider than domain knowledge in retain dataset.
In the WHP dataset, the sequential tasks involve unlearning data from several books in the Harry Potter series. Due to the strong interconnections between the data across these tasks, unlearning earlier tasks inevitably impacts subsequent tasks, leading to pronounced cascading degradation.
For baseline methods, model utility declines largely after completing the unlearning. 
In contrast, our method effectively mitigates the cascading degradation issue, demonstrating improved preservation of model utility in continual unlearning scenarios.

\subsection{Evaluation of TRAVIS Dataset Sensitivity}
The TRAVIS dataset consists of a wide variety of topics, enabling a comprehensive evaluation of the model’s overall utility. As the dataset is generated by the LLM, it provides an accurate baseline for evaluating the model's original utility. To investigate the sensitivity of the TRAVIS dataset to variations in model performance, we conducted a synthetic experiment by introducing a small amount of random Gaussian noise to the model parameters to simulate destruction of model utility. We compared the Rouge scores of the TRAVIS dataset against those of the TOFU and WHP datasets to assess their relative sensitivity to model utility changes. As presented in Table~\ref{tab:travis_sensitivity}, the TRAVIS dataset is the most sensitive to changes in model utility. Notably, the TRAVIS Rouge score exhibits a decline when the noise standard deviation reaches 0.5\%, whereas the TOFU and WHP datasets remain largely unaffected until higher noise levels.

\begin{table}[t]
\centering
\caption{Rouge Score Sensitivity to Gaussian Noise in Model Parameters.}
\label{tab:travis_sensitivity}
\begin{tabular}{c|c|c|c}
\toprule
Noise Std(\%) & TRAVIS & TOFU & WHP \\
\midrule
\rowcolor{gray!20} Original   & 0.7643 & 0.8965 & 0.6940 \\
\hline
0.1\% & 0.7643 & 0.8965 & 0.6940 \\
0.5\% & 0.7576 & 0.8965 & 0.6940 \\
1.0\%   & 0.7425 & 0.8965 & 0.6940 \\
2.0\%   & 0.7320 & 0.8958 & 0.6940 \\
\bottomrule
\end{tabular}
\end{table}

\section{Conclusion}
In this study, we explore the challenges of continual unlearning in LLMs, which is a practical and complicated setting. 
Specifically, the issues of accumulative decline and cascading degradation exacerbate the utility deterioration problem for LLM unlearning methods.
To mitigate these issues, we propose~\method, a method that minimizes adjustments to model parameters by identifying the model parameters that are crucial to the knowledge representation of each task and tuning these parameters with adaptive intensity.
The proposed method is built on the task vector framework with three novel modules applied in the fine-tuning process, dynamically localizing model parameters and adaptively refining gradients.
Besides, we construct an evaluation corpus, \texttt{TRAVIS}, designed to test the overall utility of LLMs comprehensively.
The experimental results demonstrate that the proposed method effectively preserves the utility of LLMs while accomplishing sufficient unlearning, outperforming various baseline methods.

\section*{Acknowledgments}
This work is funded by the National Science and Technology Major Project (No. 2022ZD0114903) and the Natural Science Fundation of China (NSFC. No. 62176132). Sen Cui would like to acknowledge the financial support received from Shuimu Tsinghua scholar program. Qizhou Wang and Bo Han were supported by the NSFC General Program No. 62376235, GDST Basic Research Fund Nos. 2022A1515011652 and 2024A1515012399.
This research was partially supported by Basic Research Grant (Super AI) of Institute for AI and Beyond of the University of Tokyo.

\section*{Impact Statement}
We propose a novel method for continual unlearning in large language models that is designated to protect personal privacy and copyright.
It can also be utilized to prevent harmful responses from LLMs.
For similar reasons, the proposed method also does not involve storing training data.
The constructed corpus \texttt{TRAVIS} is generated by the Phi model and the Llama2 model, only presenting general knowledge.
It is manually filtered to guarantee no private or sensitive data is included.

\nocite{langley00}

\bibliography{example_paper}
\bibliographystyle{icml2025}

\newpage
\appendix
\onecolumn

\section{Related Work}
Machine unlearning studies the problem of efficiently removing specific data from machine learning models, ensuring compliance with privacy regulations like GDPR~\cite{DBLP:conf/sp/CaoY15, wang2025gru, DBLP:journals/sncs/ZhangNIS23, DBLP:conf/nips/NiuPSMS16, DBLP:journals/corr/abs-2410-03124}.
Machine unlearning methods are generally categorized into exact unlearning~\cite{DBLP:conf/colt/UllahM0RA21,DBLP:conf/icml/BrophyL21} and approximate unlearning. The latter modifies the model or dataset to simulate the removal effect without full retraining~\cite{DBLP:conf/icml/WuDD20, DBLP:conf/eccv/GolatkarAS20, DBLP:conf/cvpr/GolatkarARPS21, DBLP:conf/aaai/GravesNG21}.
There is a line of work that employs Newton's method to remove the influence of target data involving calculating second-order derivatives of model parameters~\cite{koh2017understanding, guo2019certified, DBLP:conf/aistats/IzzoSCZ21}.
However, the approach is computationally prohibitive for LLMs.
In addition, precisely evaluating LLM unlearning algorithms, as is commonly done in the field of machine unlearning, is often impractical because obtaining a fully retrained model for comparison is rarely feasible~\cite{DBLP:journals/corr/abs-2410-03523}.

Among the studies on LLM unlearning, some focus on establishing evaluation metric~\cite{DBLP:journals/corr/abs-2406-09179, wang2025rethinking, yang2025exploring, DBLP:journals/corr/abs-2402-16835, DBLP:journals/corr/abs-2410-08109,DBLP:journals/corr/abs-2410-15153}, while others emphasize the design of effective algorithms.
With trivial unlearning methods, the process often leads to excessive unlearning, where the performance of the model on non-target data is unintentionally degraded.
The primary objective of LLM unlearning is to preserve the model utility on normal data while effectively unlearning the target information.
NPO is proposed to mitigate the issue of unbounded optimization loss of GA~\cite{DBLP:journals/corr/abs-2404-05868}.
\citet{DBLP:journals/corr/abs-2310-02238} suggest replacing sensitive keywords with generic terms, enabling the model to produce generic predictions through reinforcement bootstrapping.
The constructed generic outputs enable unlearning by fine-tuning the LLM model on them.
\citet{DBLP:journals/corr/abs-2406-08607} propose ULD, which reverses the optimization direction during the training of an assistant model.
SKU incorporates an object function that embeds robustness and utility retention into task vector training~\cite{DBLP:conf/acl/0010DT0024}, offering a more efficient approach to unlearning.
WAGLE seeks to locate the specific LLM parameters that are most critical for both unlearning and retaining objectives and fine-tuning exclusively on these parameters~\cite{DBLP:journals/corr/abs-2410-17509}.
Several studies have also examined the impact of unlearning techniques on various modules within transformer models~\cite{DBLP:conf/emnlp/HongZHZ0Y24,DBLP:journals/corr/abs-2406-11614}.
Despite the substantial body of research on LLM unlearning, the practical challenge of continual unlearning remains underexplored and warrants further investigation.

\section{Convergence of GA and the task vector method}
\label{sec:convergence_tv_ga}

To compare the convergence of GA and the task vector method, we consider a standard binary classification problem with training data $\{(x_i, y_i)\}_{i=1}^n$, where $x\in\mathbb{R}^{d_l}$ and $y\in\{0,1\}$. A logistic model is used to predict the probabilities $\pi(y=1|x,\theta)=\sigma(\langle x,\theta\rangle)$, where $\sigma$ represents the sigmoid function~\cite{schmidt2017cpsc}.

\subsection{Convergence of the task vector method}
\begin{theorem}
Using the task vector algorithm with a learning rate $0<\eta<\frac{2}{L}$ to optimize the aforementioned logistic regression problem can guarantee convergence. Let $\theta^t$ denotes the parameters after the $t$-th iteration of optimizing. The loss function $f_\text{TV}(\theta^t)$ strictly decreases:
\begin{equation}
    f_\text{TV}(\theta^{t+1})<f_\text{TV}(\theta^{t}),~t\geq0,
\end{equation}
and converges to its optimal value:
\begin{equation}
    \lim_{t\rightarrow\infty}f_\text{TV}(\theta^t)=f_\text{TV}(\theta^*)
\end{equation}
\end{theorem}
\textbf{Proof}. The fine-tuning process of the task vector method utilizes the gradient descent algorithm with the following loss function:
\begin{equation}
    f_\text{TV}(\theta^t)=-\sum_{i}y_i\log\sigma(\langle x,\theta^t\rangle)-(1-y_i)\log\sigma(-\langle x,\theta^t\rangle).
\end{equation}
The function $f_\text{TV}$ is L-smooth. Thus the following inequality holds:
\begin{equation}
    f_\text{TV}(\theta^{t+1})\leq f_\text{TV}(\theta^{t})+\nabla f_\text{TV}(\theta^{t})^\top(\theta^{t+1}-\theta^{t})+\frac{L}{2}||\theta^{t+1}-\theta^{t}||^2.
\end{equation}
Substituting $\theta^{t+1}-\theta^{t}=-\eta \nabla f_\text{TV}(\theta^{t})$ yields:
\begin{equation}
    f_\text{TV}(\theta^{t+1})\leq f_\text{TV}(\theta^{t})-\eta\left(1-\frac{L\eta}{2}\right)||\nabla f_\text{TV}(\theta^{t})||^2.
\end{equation}
With $0<\eta<\frac{2}{L}$, we get a strict decrease in $f_\text{TV}(\theta^{t})$:
\begin{equation}
    f_\text{TV}(\theta^{t+1})<f_\text{TV}(\theta^{t})
\end{equation}
With lower bound $f_\text{TV}(\theta)\geq 0$, the loss function converges to its optimal value:
\begin{equation}
    \lim_{t\rightarrow\infty}f_\text{TV}(\theta^t)=f_\text{TV}(\theta^*)
\end{equation}

\subsection{Non-convergence of the GA method}
\label{sec:non_converge_GA}
\begin{theorem}
Optimizing the aforementioned logistic regression problem using the GA algorithm will result in the norm of the gradient remaining greater than a certain constant:
\begin{equation}
    ||\nabla_{\theta^t} f_{\text{GA}}(\theta^t)||\geq\sqrt{C},
\end{equation}
where $\theta^t$ denotes the parameters after the $t$-th iteration of optimizing and $C$ is a positive constant only depending on the training data $\{(x_i, y_i)\}_{i=1}^n$ and initial parameters $\theta^0$.
And the objective function $f_{\text{GA}}(\theta^t)$ strictly decreasing to negative infinity:
\begin{equation}
\begin{aligned}
&f_{\text{GA}}(\theta^{t})= f_{\text{GA}}(\theta^{0})-t\eta C,\\
&\lim_{t\rightarrow\infty}f_{\text{GA}}(\theta^{t})=-\infty.
\end{aligned}
\end{equation}
\end{theorem}
\textbf{Proof}.
The loss function of GA is:
\begin{equation}
    f_{\text{GA}}(\theta^t)=\sum_{i}y_i\log\sigma(\langle x,\theta^t\rangle)+(1-y_i)\log\sigma(-\langle x,\theta^t\rangle)
\end{equation}
The gradient with respect to the parameters $\theta^t$ is:
\begin{equation}
\begin{aligned}
    \nabla_{\theta^t}& f_{\text{GA}}(\theta^t)=\sum_i (2y_i-1)(1-\pi_i^t)x_i,\\
    \text{where}&~\pi_i^t=\sigma(\langle x,\theta\rangle)~\text{when}~y_i=1,\\
    &\pi_i^t=\sigma(-\langle x,\theta\rangle)~\text{when}~y_i=0.
\end{aligned}
\end{equation}
We define $m_i^t=(2y_i-1)(1-\pi_i^t),~\gamma_{i,j}=\langle x_i,x_j\rangle$, and $\gamma_i$ denotes $(\gamma_{i,1}, \gamma_{i,2},...,\gamma_{i,n})$. Besides, we define $q_i^t=\langle m^t, \gamma_i\rangle$.

Inspired by~\citet{DBLP:journals/corr/abs-2404-05868}, we use the induction method to proceed.

\textbf{Case 1}:~$t=0$. Since $||x_i||_2\cdot||\theta^0||_2$ is a finite value, we can derive:
\begin{equation}
    \begin{aligned}
        C_1\leq m_i^0\leq1~\text{when}~y_i=1,\\
        -1\leq m_i^0\leq-C_2~\text{when}~y_i=0,
    \end{aligned}
\end{equation}
where $C_1$ and $C_2$ are two positive constants. We assume there exists a constant $C_0$ such that $\max_{i\neq j}|\gamma_{i,j}|\leq \frac{C_0}{n}$, and satisfying:
\begin{equation}
    \left|\sum_{i\neq j}m_j^0\gamma_{i,j}\right|\leq\frac{\gamma_{i,i}|m_i^0|}{2}.
\end{equation}
We obtain:
\begin{equation}
\begin{aligned}
    q_i^0=\langle m^0, \gamma_i\rangle\geq \frac{|m_i|\gamma_{i,i}}{2}>0~\text{when}~y_i=1,\\
    q_i^0=\langle m^0, \gamma_i\rangle\leq -\frac{|m_i|\gamma_{i,i}}{2}<0~\text{when}~y_i=0.
\end{aligned}
\end{equation}
We set $C_3=\min(C_1,C_2)$ and $C_4=\frac{C_3\gamma_{i,i}}{2}$, then we have:
\begin{equation}
\begin{aligned}
    q_i^0=\langle m^0, \gamma_i\rangle\geq C_4>0~\text{when}~y_i=1,\\
    q_i^0=\langle m^0, \gamma_i\rangle\leq -C_4<0~\text{when}~y_i=0.
\end{aligned}
\end{equation}
Then we derive $m^1_i$:
\begin{equation}
\begin{aligned}
    m^1_i=(2y_i-1)(1-\pi_i^1)=(2y_i-1)(1-\sigma((2y_i-1)\langle x_i,\theta^1\rangle)).
\end{aligned}
\end{equation}
Substracting $\theta^1=\theta^0-\eta\nabla_{\theta^t}f_{\text{GA}}(\theta^0)=\theta^0-\eta\sum_i m_i^0x_i$, we obtain:
\begin{equation}
    m_i^1=(2y_i-1)\left(1-\sigma\left((2y_i-1)\langle x_i,\theta^0\rangle-\eta(2y_i-1)q_i^0\right)\right).
\end{equation}
Comparing it with $m_i^0=(2y_i-1)(1-\sigma((2y_i-1)\langle x_i,\theta^0\rangle))$, we can derive:
\begin{equation}
\begin{aligned}
    m_i^1\geq m_i^0~\text{when}~y_i=1\\
    m_i^1\leq m_i^0~\text{when}~y_i=0.
\end{aligned}
\end{equation}

\textbf{Case 2}:~$t=K>0$. We suppose:
\begin{equation}
\begin{aligned}
    q_i^t\geq C_4>0~\text{when}~y_i=1,\\
    q_i^t\leq -C_4<0~\text{when}~y_i=0,
\end{aligned}
\end{equation}
for $0\leq t\leq K$. And we suppose:
\begin{equation}
\begin{aligned}
    m_i^{t+1}\geq m_i^t\geq C_1~\text{when}~y_i=1\\
    m_i^{t+1}\leq m_i^t\leq -C_2~\text{when}~y_i=0,
\end{aligned}
\end{equation}
for $0\leq t\leq K-1$. With this condition, we can derive the same conclusion for $q_i^{K+1}$ as in Case 1:
\begin{equation}
\begin{aligned}
    q_i^{K+1}\geq C_4>0~\text{when}~y_i=1,\\
    q_i^{K+1}\leq -C_4<0~\text{when}~y_i=0,
\end{aligned}
\end{equation}
and the same conclusion for $m_i^{K+1}$:
\begin{equation}
\begin{aligned}
    m_i^{K+1}\geq m_i^K\geq C_1~\text{when}~y_i=1\\
    m_i^{K+1}\leq m_i^K\leq -C_2~\text{when}~y_i=0.
\end{aligned}
\end{equation}

From above, we prove that $|q_i^t|\geq C_4$ and $(2y_i-1)m_i^{t+1}>(2y_i-1)m_i^{t}$ for all $t\geq0$. We can further derive that:
\begin{equation}
    f_{\text{GA}}(\theta^{t+1})\leq f_{\text{GA}}(\theta^{t}),
\end{equation}
and
\begin{equation}
    ||\nabla_{\theta^t} f_{\text{GA}}(\theta^t)||_{X^\top X}=\sqrt{\nabla_{\theta^t} f_{\text{GA}}(\theta^t)^\top X^\top X \nabla_{\theta^t} f_{\text{GA}}(\theta^t)}=\sqrt{(q^t)^\top q^t}\geq C_4\sqrt{n}.
\end{equation}
By incorporating the condition $||\nabla_{\theta^t} f_{\text{GA}}(\theta^t)||_{X^\top X}\leq\sqrt{\lambda_{\text{max}}}||\nabla_{\theta^t} f_{\text{GA}}(\theta^t)||_2$, we can obtain:
\begin{equation}
    ||\nabla_{\theta^t} f_{\text{GA}}(\theta^t)||_2\geq \frac{C_4\sqrt{n}}{\sqrt{\lambda_{\text{max}}}}=\sqrt{C},
\end{equation}
where $C$ is a positive constant only depending on the training data $\{(x_i, y_i)\}_{i=1}^n$ and initial parameters $\theta^0$.

Furthermore, since $f_{\text{GA}}(\theta)$ is a concave function, it satisfies the following inequality:
\begin{equation}
\begin{aligned}    
    &f_{\text{GA}}(\theta^{t+1})\leq f_{\text{GA}}(\theta^{t})+\nabla_{\theta^t} f_{\text{GA}}(\theta^t)^\top(\theta^{t+1}-\theta^{t})=f_{\text{GA}}(\theta^{t})-\eta||\nabla_{\theta^t} f_{\text{GA}}(\theta^t)||_2^2,
\end{aligned}
\end{equation}
$\text{then}~f_{\text{GA}}(\theta^{t+1})\leq f_{\text{GA}}(\theta^{t})-\eta C$, and we can obtain:
\begin{equation}
    \lim_{t\rightarrow\infty}f_{\text{GA}}(\theta^{t})=\lim_{t\rightarrow\infty} f_{\text{GA}}(\theta^{0})-t\eta C=-\infty
\end{equation}

\section{Proof of Proposition~\ref{prop:cul}}
\label{sec:app_proof_cul}

As defined for the GA, the loss function and its gradient are given as follows:
\begin{equation}
\begin{aligned}
    f_{\text{GA}} (\theta) &=\sum_{i}y_i\log\sigma(\langle x,\theta\rangle)+(1-y_i)\log\sigma(-\langle x,\theta\rangle)\\
    \nabla_{\theta} f_{\text{GA}}(\theta)&=\sum_i (2y_i-1)(1-\pi_i)x_i,
\end{aligned}    
\end{equation}
where $\pi_i$ follows the definition in Appendix~\ref{sec:non_converge_GA}, as do $q_i$, $m_i$ and $\gamma_{i,j}$.

\textbf{Independently training on $D^s$.}

In this scenario, the model is directly trained on $D^s$ starting from the initial parameter $\theta^0$ for $T$ iterations with a learning rate $\eta$, resulting in the updated parameters $\theta_\text{IUL}$:
\begin{equation}
\begin{aligned}
\theta_\text{IUL}^{t+1}=\theta_\text{IUL}^{t}-\eta \sum_i m_{\text{IUL},i}^t x^s_i,\\
\langle\theta_\text{IUL}^{T}, x^s_i\rangle=\langle\theta^{0},x_i^s\rangle-\eta \sum_t q_{\text{IUL},i}^t.
\end{aligned}
\end{equation}

As in Appendix~\ref{sec:non_converge_GA}, we can prove that:
\begin{equation}
\begin{aligned}
    q_{\text{IUL},i}^t\geq C_4>0~\text{when}~y_i=1,\\
    q_{\text{IUL},i}^t\leq -C_4<0~\text{when}~y_i=0,
\end{aligned}
\end{equation}

\textbf{Continuously training on $D^f$ and $D^s$.}

In this scenario, the model is first trained on $D^f$ starting from $\theta^0$, yielding parameters $\theta_f$:
\begin{equation}
\theta_\text{s}^{t+1}=\theta_\text{s}^{t}-\eta \sum_i m_{\text{s},i}^t x^s_i
\end{equation}

The training on $D^f$ affects the model predictions on $D^s$:
\begin{equation}
\langle\theta_f^{t+1}, x^s_i\rangle=\langle\theta_f^{t}, x^s_i\rangle-\eta \sum_j m_{f,i}^t\langle x_{f,j},x_{s,i}\rangle
\end{equation}

Because $\max_{i\neq j}|\langle x^f_j,x^s_i\rangle|<k\langle x^f_i,x^s_i\rangle$, we assume there exists a constant $C_0$ such that $\max_{i\neq j}|\langle x^f_j,x^s_i\rangle|\leq \frac{C_0}{n}$, and satisfying:
\begin{equation}
    \left|\sum_{i\neq j}m_{f,i}^t|\langle x^f_j,x^s_i\rangle|\right|\leq\frac{\langle x^f_i,x^s_i\rangle|m_{f,i}^t|}{2}.
\end{equation}

Therefore, we can obtain:
\begin{equation}
\begin{aligned}
    \langle\theta_f^{t+1}, x^s_i\rangle<\langle\theta_f^{t}, x^s_i\rangle-\frac{\langle x^f_i,x^s_i\rangle|m_{f,i}^t|}{2}~\text{when}~y_i=1,\\
    \langle\theta_f^{t+1}, x^s_i\rangle>\langle\theta_f^{t}, x^s_i\rangle+\frac{\langle x^f_i,x^s_i\rangle|m_{f,i}^t|}{2}~\text{when}~y_i=0.
\end{aligned}
\end{equation}

Thus the effect of training on $D^f$ demonstrates as:
\begin{equation}
\begin{aligned}
    \langle\theta_f, x^s_i\rangle<\langle\theta^{0}, x^s_i\rangle-C_1T~\text{when}~y_i=1,\\
    \langle\theta_f, x^s_i\rangle>\langle\theta^{0}, x^s_i\rangle+C_1T~\text{when}~y_i=0.
\label{eq:effect_first}
\end{aligned}
\end{equation}

Then the model starts training on $D^s$ with initial parameters being $\theta_f$, yielding updated parameters $\theta_{\text{CUL}}$:
\begin{equation}
\begin{aligned}
\theta_\text{CUL}^{t+1}=\theta_\text{CUL}^{t}-\eta \sum_i m_{\text{CUL},i}^t x^s_i,\\
\langle\theta_\text{CUL}^{T}, x^s_i\rangle=\langle\theta_{f},x_i^s\rangle-\eta \sum_t q_{\text{CUL},i}^t.
\end{aligned}
\end{equation}

Since $\theta_\text{CUL}^{0}=\theta_f$ the inequality in Equation~\ref{eq:effect_first}, we can derive:
\begin{equation}
\begin{aligned}
    q_{\text{CUL},i}^0>q_{\text{IUL},i}^0+C_2~\text{when}~y_i=1,\\
    q_{\text{CUL},i}^0<q_{\text{IUL},i}^0-C_2~\text{when}~y_i=0.
\end{aligned}
\end{equation}
Using the induction method as in Appendix~\ref{sec:non_converge_GA}, we can prove that:
\begin{equation}
\begin{aligned}
    q_{\text{CUL},i}^t>q_{\text{IUL},i}^t+C_2>0~\text{when}~y_i=1,\\
    q_{\text{CUL},i}^t<q_{\text{IUL},i}^t-C_2<0~\text{when}~y_i=0.
\end{aligned}
\end{equation}
Therefore, $|\sum_t q_{\text{CUL},i}^t|>|\sum_t q_{\text{IUL},i}^t|+TC_2$.

Finally, we derive the conclusion:
\begin{equation}
    ||\theta_\text{CUL}-\theta_f||_{X^\top X}=||\sum_t q_{\text{CUL}}^t||_2>||\sum_t q_{\text{CUL}}^t||_2+TC\sqrt{n},
\end{equation}
therefore,
\begin{equation}
    ||\theta_\text{CUL}-\theta_f||_{X^\top X}>||\theta_\text{IUL}-\theta^0||_{X^\top X}+TC\sqrt{n}.
\end{equation}

In this proposition, we consider a two-task scenario for simplicity. When there are multiple tasks, we can also similarly derive such conclusions:
\begin{corollary}
For multiple tasks, the conclusion of Proposition~\ref{prop:cul} holds. Suppose there are $T$ preceding datasets $\{D^t\}_{t=1}^T$ and a current dataset $D^s$. If one of the preceding datasets is correlated with $D^s$, satisfying: $\max_{i\neq j}|\langle x_j,x^s_i\rangle|<k\langle x_i,x^s_i\rangle$, then the previous unlearning results in larger changes of parameters in the unlearning of $D^s$:
\begin{equation}
    ||\bigtriangleup\theta_\text{CUL}||_{X^\top X}>||\bigtriangleup\theta_\text{IUL}||_{X^\top X}+TC\sqrt{n}.
\end{equation}
\end{corollary}

\section{Derivation of Equation~\ref{eq:mt_it}}
\label{sec:derivation_eq_mt_it}
According to the chain rule, the gradient of Equation~\ref{eq:mt_obj} with respect to $m^t$ is given by:
\begin{equation}
    -\lambda\nabla_{\theta_\text{ft}^{t,k+1}}\mathcal{L}_{CE}(\theta^{t-1}-\lambda(\theta_{\text{ft}}^{t,k+1}-\theta^{0}), D_r)\nabla_{m^t}\theta_\text{ft}^{t,k+1}+\mu M\nabla_{m^t}\mathds{1}_{m^t\geq\eta}
\end{equation}
For computational efficiency, we approximate the gradient of $\mathcal{L}_{CE}$ on $D_r$ with $\nabla_{\theta^{t-1}}\mathcal{L}_{CE}(\theta^{t-1}, D_r)$. And $\nabla_{m^t}\theta_\text{ft}^{t,k+1}$ is derived by substituting into Equation~\ref{eq:theta}:
\begin{equation}
    \nabla_{m^t}\theta_\text{ft}^{t,k+1}=-\hat{\alpha}^t\odot\mathcal{L}_{ESL}(\theta_{\text{ft}}^{t,k}, D_u^t)\odot\nabla_{m^t}\mathds{1}_{m^t\geq\eta}
\end{equation}
For simplicity, we omit $\hat{\alpha}^t$ and approximate $\nabla_{m^t}\mathds{1}_{m^t\geq\eta}$ using straight-through estimation, which consists in taking this derivative equal to the identity~\cite{DBLP:conf/nips/OswaldZKSCZS21}. Then we obtain the gradient of Equation~\ref{eq:mt_obj} with respect to $m^t$:
\begin{equation}
    -\lambda\mathcal{L}_{ESL}(\theta_{\text{ft}}^{t,k}, D_u^t)\odot\nabla_{\theta^{t-1}}\mathcal{L}_{CE}(\theta^{t-1}, D_r)+\mu M
\end{equation}

The updating of $m^t$ is given by substracting the gradient:
\begin{equation}
    m^t\leftarrow m^t+\lambda\mathcal{L}_{ESL}(\theta_{\text{ft}}^{t,k}, D_u^t)\odot\nabla_{\theta^{t-1}}\mathcal{L}_{CE}(\theta^{t-1}, D_r)-\mu M
\end{equation}

\section{Implementation details}
\label{sec:implementation}

\subsection{Optimization}
The Adam optimizer is employed for the training of all methods.
We employ the learning rate of 3e-5 for the TOFU dataset and 3e-6 for the MUSE News and WHP datasets.
Most baselines easily suffer from the issue of over-unlearning while conducting second or third tasks with constant learning rate.
Thus, we apply learning rate tuning, decaying the learning rate by 50\% after each unlearning task.
For experiments in Figure~\ref{fig:cascadingF} and Figure~\ref{fig:continual_onetime_unlearning_compare} which require alignment of unlearning performance, learning rates of different methods are tuned to achieve aimed unlearning performance.
The hyperparameters $s$ and $\lambda$ of our method are set as 10 and 0.8.
$\mu$ is set as the deviation of the unlearning gradients $G_u$.
$m^t$ is initialized with zero vectors in the beginning of each training.

Weight decay ratio is set as 0.01 for all experiments.
For TOFU dataset, batch size is set as 32 and models are trained for 5 epochs.
For MUSE News dataset, batch size is 6 and number of epochs is 2.
For WHP dataset, batch size and epoch number are 6 and 10.

\subsection{Tuning of gradient mask threshold $\eta$}
\label{sec:tuning_of_eta}
The threshold $\eta$ determines the proportion of model parameters that are activated.
The threshold $\eta$ is determined by the percentiles of the underlying vector $m^t$, and it increases gradually throughout the training process.
Specifically, 100\% parameters are activated in the beginning of each training, while 30\% parameters are activated in the end.
With this approach, parameters with smaller $m^t$ values undergo minor fine-tuning in the early stages of training, while parameters with larger $m^t$ values receive more extensive fine-tuning over additional steps.
Compared to setting a large threshold $\eta$ in the beginning of training, this method balances the training of all model parameters, preventing excessive adjustments to a small subset of activated parameters to achieve the unlearning objective.

Calculating the percentiles of $m^t$ can be computationally expensive for models with a large number of parameters since it involves sorting the parameters.
To address this, we employ an efficient estimation method.
Calculating the maximum and minimum values of a list has a computational complexity of $O(n)$.
By identifying the top 15\% largest and smallest values, we can mitigate the influence of outliers.
The estimated quantiles are then obtained through interpolation between these extreme values after excluding outliers.

\subsection{Discussions regarding parameters masking in LLMs unlearning}
We argue that an efficient way to preserve the utility of LLM during the unlearning process is to minimize changes to its parameters.
In contrast to methods that directly calculate the mask using gradient magnitude, our proposed dynamic gradient sparsity method in Section~\ref{sec:dynamic_gradient} learns the mask values from the data.
And the mask is dynamically tuned along with the training process.
In this way, each model parameter receives an appropriate training range: parameters with a relatively small impact on the training objective undergo minor adjustments in the early stages, after which they are masked, while parameters with a larger impact are trained more extensively. This approach allows for appropriate training of the model as a whole, avoiding large adjustments to all parameters or drastic changes to a small subset.
Besides, the proposed method leverages the underlying vector $m$ to accumulate the influence of data on the mask, thereby avoiding biases introduced by extreme values during a single training iteration.

\subsection{Improving Memory Efficiency}
\label{sec:memory_efficiency}	
Although effective, the iterative approach in Equation~\ref{eq:mt_it} requires storing $G_r=\nabla_{\theta^{t-1}}\mathcal{L}_{CE}(\theta^{t-1}, D_r)$ during the fine-tuning process of the $t$-th task. To reduce memory usage, we propose an alternative method.%

To reduce memory overhead during training, $G_r$ can be replaced with a lower-precision approximation, $\hat{G_r}$. For elements with absolute values exceeding a certain threshold, which have a significant impact on the retain set, their signs $+1$ or $-1$ are preserved, while the remaining elements are set to $0$.
During the iteration of $m^t$, calculation of $G_u\odot G_r$ is estimated with:
\begin{equation}
    G_u\odot(\hat{G_r}==0)+a\cdot G_u\odot(\hat{G_r}==1)-a\cdot G_u\odot(\hat{G_r}==-1),
\end{equation}
where $a$ is a large positive constant.
This method preserves both the signs and magnitude information of $G_r$ while significantly reducing memory consumption.

\section{Experimental Setup}
\subsection{Dataset}

\textbf{TOFU}~\cite{DBLP:journals/corr/abs-2401-06121} dataset consists of information about 200 fictional authors presented in a question-answer pair format. Specific authors' information was designated as unlearning data, while the remaining authors' data were retained to assist in training and testing. Besides, there are also question-answer pairs about \textit{Real Authors} and \textit{World Facts}. We grouped the information of four authors into one unlearning task, resulting in five or more continual unlearning tasks.

\textbf{MUSE News}~\cite{DBLP:journals/corr/abs-2407-06460} comprises BBC news articles published after August 2023. This dataset includes four predefined unlearning subsets that serve as continual unlearning tasks. Additionally, it contains a retain set that does not overlap with the unlearning sets.

\textbf{WHP} (Who's Harry Potter)~\cite{DBLP:journals/corr/abs-2310-02238,DBLP:journals/corr/abs-2407-06460} involves the unlearning of original data from the Harry Potter series. In contrast, retain data consists of knowledge related to the books, which aligns with real-world requirements for copyright protection. We treated the original texts of three Harry Potter books as three separate unlearning tasks.
To evaluate the performance of algorithms in scenarios involving unrelated continual tasks, we constructed a sequence of continual unlearning tasks by interleaving multiple unlearning tasks from the MUSE News and WHP datasets.

\subsection{Metrics}
\textbf{Rouge-L for unlearning (F-Rouge)} measures the similarity between the text generated by the model and the correct text. Specifically, given a question or the beginning of the unlearning text, the Rouge-L recall is computed by comparing the output of models with the correct answer or remaining text. A lower Rouge-L value indicates better unlearning performance~\cite{DBLP:journals/corr/abs-2401-06121, DBLP:journals/corr/abs-2407-06460}.

\textbf{Probability (F-Prob)} evaluates the conditional probability $P(\text{answer}|\text{question})$ of the model outputs in the TOFU dataset. To ensure comparability, this probability is normalized by the length of the output text~\cite{DBLP:journals/corr/abs-2401-06121}.

\textbf{Membership inference attack (MIA)} uses the state-of-the-art Min-K\% Prob method for LLMs~\cite{DBLP:conf/iclr/ShiAXHLB0Z24} to detect whether data in the unlearning set remains part of the training data for the post-unlearning model. A lower probability of the unlearning set belonging to the training data reflects a more effective unlearning process.

\textbf{Forget Quality (FQ)} quantifies the difference in prediction distributions on the forget set between the post-unlearning model and a retrained model (fine-tuned on retrained data without exposure to the forget data). This metric assesses the degree to which the unlearning algorithm eliminates traces of the forget set in the model~\cite{DBLP:journals/corr/abs-2401-06121}.

\textbf{R-Rouge} and \textbf{R-Prob} are also applied to the retain set or holdout set to evaluate the impact of the unlearning algorithm on the overall model utility. Higher values for these metrics indicate less performance degradation.

\textbf{T-Rouge} calculates the Rouge-L recall of completing sentences in the proposed TRAVIS dataset. Testing on inferred  training data provides a more accurate assessment of the impact of unlearning algorithms on the pre-existing knowledge of the model. Additionally, TRAVIS is generated without thematic constraints, resulting in a broad coverage of content, which enables a more comprehensive evaluation of performance.

\textbf{Model Utility (MU)} represents the harmonic mean of the above metrics calculated across multiple retain sets in the TOFU dataset~\cite{DBLP:journals/corr/abs-2401-06121}, including TRAVIS dataset. It provides an aggregate measure of the model utility.

\subsection{Baseline methods}
\textbf{GA}~\cite{DBLP:journals/corr/abs-2310-10683} unlearns target data by optimizing in the opposite direction of gradient descent. This is achieved by minimizing the negative cross-entropy loss of predicting the next token on the unlearning dataset:
\begin{equation}
	\min_{\theta} -\frac{1}{n_u}\sum_{x\in D_u}\sum_{l}-\log{\pi_{\theta}(x_{[l+1]}|x_{[:l]})}.
\end{equation}

\textbf{NPO}~\cite{DBLP:journals/corr/abs-2404-05868} is also a gradient ascent-based approach. Compared to GA, the gradient in NPO is scaled by a regularization term that gradually diminishes over the course of training, alleviating the problem of over-unlearning to some extent. The loss objective of NPO is:
\begin{equation}
    \frac{2}{\beta}\left[\log\left(1+\left(\frac{\sum_{l}\pi_\theta(x_{[l+1]}|x_{[:l]})}{\sum_{l}\pi_\text{ref}(x_{[l+1]}|x_{[:l]})}\right)^\beta\right)\right]
\end{equation}

\textbf{Task Vectors} (\textbf{TV})~\cite{DBLP:conf/iclr/IlharcoRWSHF23} method manipulates the ability of a pre-trained model to address downstream tasks through fine-tuning and the arithmetic operation of model parameters.
When applied to LLM unlearning, the method enables the selective removal of knowledge related to a specific dataset $D_u$.
Specifically, a pre-trained model $\pi_{\theta}$ is fine-tuned on the unlearning dataset $D_u$, resulting in a model $\pi_{\theta_{\text{ft}}}$ with enhanced knowledge of $D_u$. The difference between the fine-tuned parameters and the original parameters, $\theta_{\text{ft}}-\theta$, constitutes the task vector for $D_u$.
By subtracting this task vector from the original parameters $\theta$, a modified model is obtained that has unlearned the knowledge from $D_u$.

\textbf{DPO} is a preference optimization approach.
For a preference dataset, DPO encourages the model to produce one output over the other between a pair.
In the unlearning scenario, it encourages the model to produce responses $y_w$ like "I don't know" instead of generating original sensitive text $y_l$~\cite{DBLP:journals/corr/abs-2404-05868}.

\textbf{SKU}~\cite{DBLP:conf/acl/0010DT0024} is a method based on task vectors.
In the fine-tuning process, SKU not only reinforces the target knowledge into the model, but also eliminates knowledge of retain set from the model, incorporating a utility preservation objective.
Besides, it includes a robustness objective during the fine-tuning process.

\textbf{EUL} incorporates lightweight unlearning layers into the model to enable efficient unlearning. This method explicitly considers scenarios involving multiple unlearning tasks~\cite{DBLP:conf/emnlp/ChenY23}.

\textbf{SO-PO} and \textbf{SO-NPO} introduce a second-order optimization framework for LLM unlearning, leveraging Hessian matrix approximations to address the challenges of applying Newton's method to LLMs~\cite{DBLP:conf/emnlp/JiaZZLRDK024}.

\textbf{WAGLE} identifies model parameters that benefit both unlearning and retaining objectives before training begins, optimizing only these parameters during the unlearning process. This approach is utilized to minimize the impact of unlearning on model utility~\cite{DBLP:journals/corr/abs-2410-17509}.

When applying these methods in continual unlearning scenarios, the learning rate is tuned progressively to mitigate catastrophic collapse as tasks continue.

\section{TRAVIS Dataset}

To construct a high-quality text corpus, we adopted the methodology of Membership Inference Attack (MIA)~\cite{DBLP:conf/csfw/YeomGFJ18} and applied it to two pre-trained language models, Phi-1.5 and Llama2-7b.
The approach began with the generation of 5,000 sentences using the models.
Subsequently, the MIA technique was employed to identify 250 candidate sentences with the highest probabilities from the generated set~\cite{DBLP:conf/uss/CarliniTWJHLRBS21}.
After deduplication and filtering based on sentence similarity, the candidate set was further refined. More than 150 high-quality sentences were manually selected to ensure both diversity and authenticity in the dataset.

To enhance the variety of sentence prefixes and enable the models to process a broader range of contextual information, we incorporated a large collection of prefix data sourced from the internet following the work of \citet{DBLP:conf/uss/CarliniTWJHLRBS21}. Specifically, we extracted samples from a subset of the Common Crawl dataset, cleaned the data by removing HTML tags and JavaScript code, and performed deduplication, resulting in approximately 50MB of clean text data. From this corpus, we randomly selected 5 to 10 context tokens and employed a Top-n sampling strategy to generate sequences based on the model’s probability distribution. The sequences with the highest likelihood were selected, and sentence lengths were capped at a maximum of 1024 tokens to ensure relevance and appropriate length.

The MIA methodology was applied next, using the perplexity of each sentence as a measure of training data likelihood. Perplexity was calculated with the following formula:
\begin{equation}
	\mathcal{P} = \exp\left( -\frac{1}{n} \sum_{i=1}^n \log f_\theta(x_i \mid x_1, \dots, x_{i-1}) \right),
\end{equation}
where $x_1,...,x_n$ denotes a sequence of tokens. Low perplexity indicates that the model holds high confidence in completing the sequence.

However, relying solely on MIA for membership inference introduces potential risks, as the model might assign higher likelihoods to unremarkable or low-quality samples, such as repeated substrings. To address this issue, we introduced several quality control mechanisms to ensure the integrity of the generated text. Specifically, we calculated the zlib entropy of each sample, measured perplexity within a sliding window, and computed the ratio of perplexity before and after lowercasing the text to assess its complexity and information density. These metrics allowed us to filter out low-quality samples and ensure that the selected texts exhibited a high degree of linguistic fluency and semantic richness.

During the selection process, samples were ranked according to the three quality metrics, and the top 100 were chosen. After deduplication, Sentence-BERT was used to compute cosine similarity between sentences, and those with similarity scores above 0.7 were excluded. Cosine similarity was calculated by:
\begin{equation}
	\text{cos}(\mathbf{v}_1, \mathbf{v}_2) = \frac{\mathbf{v}_1 \cdot \mathbf{v}_2}{\|\mathbf{v}_1\| \|\mathbf{v}_2\|},
\end{equation}
where $\mathbf{v}_1, \mathbf{v}_2$ denote the embeddings of two sentences.
This process yielded approximately 250 diverse and high-quality sentences.
In the final manual filtering step, short, highly repetitive, or logically incoherent sentences were discarded, while longer substrings with richer semantics and more varied content were retained. The resulting dataset contains more than 150 high-quality sentences per model, meeting our established quality standards.
For dataset construction, the first 70\% of each sentence was designated as the prompt, while the remaining 30\% served as the ground truth, thus creating the final training dataset.

This dataset offers two significant advantages. First, it accurately evaluates the impact of unlearning algorithms on the original knowledge retained by models. This aligns closely with the definition of efficiency in the context of machine unlearning.
Second, the broad thematic coverage of the dataset enables evaluations of model performance across diverse contexts, ranging from everyday conversations to specialized topics. This diversity not only allows the dataset to comprehensively assess the language generation capabilities of the model, but also provides insights into its generalization ability when exposed to different background knowledge. Consequently, the dataset serves as a versatile and reliable foundation for evaluating model performance in a wide range of scenarios.

\begin{table*}[tbh]
	\centering
	\caption{Parts of examples in TRAVIS dataset (Phi Model)}
	\resizebox{\linewidth}{!}{
		\begin{tabular}{|>{\raggedright\arraybackslash}m{0.03\linewidth}|>{\raggedright\arraybackslash}m{0.9\linewidth}|} %
			\hline
			\textbf{} & \textbf{Phi Model} \\
			\hline
			\small 1 & \small The Concept and Examples Recursion is a powerful programming concept that involves a function calling itself in its definition. This allows the program to be written in a more concise and elegant way, and provides a elegant solution to problems that would otherwise require multiple lines of code. For example: def factorial(n): if n == 0: return 1 return n * factial( n-1 ) print( factroy( 10 ) ) \# Prints: 10 * 9 * 8 * 7 * 6 * 5 * 4 * 3 * 2 * 1 = 3628800\\
			\hline
			\small 2 & \small Bullying is a pervasive problem that affects many people, especially children. It can have serious long-term effects on mental health, self-esteem, and academic performance. \\
			\hline
			\small 3 & \small Financial independence is a goal that many individuals aspire to achieve, but the path to attaining it can be challenging. One of the most effective ways to build financial independence early on is through saving. \\
			\hline
			\small 4 & \small As the world continues to grapple with the effects of climate change, many are turning to creative solutions to help mitigate its impact. From art installations that draw attention to environmental issues to community-based initiatives that promote sustainability, individuals and organizations alike are stepping up to the challenge. \\
			\hline
			\small 5 & \small Reasons to do exercise regularly: 1.To maintain a healthy weight: Regular exercise helps to burn excess calories and prevent weight gain. \\
			\hline
		\end{tabular}
	}
	\label{tab:sample_6x2_centered}
\end{table*}

\begin{table*}[tbh]
	\centering
	\caption{Parts of examples in TRAVIS dataset (Llama Model)}
	\resizebox{\linewidth}{!}{
		\begin{tabular}{|>{\raggedright\arraybackslash}m{0.03\linewidth}|>{\raggedright\arraybackslash}m{0.9\linewidth}|} %
			\hline
			\textbf{} & \textbf{Llama Model} \\
			\hline
			\small 1 & \small The city of Los Angeles also offers a wide range of public services, including healthcare, education, and transportation, all of which are critical in maintaining a high quality of life for city residents. \\
			\hline
			\small 2 & \small There are a few things you can do to make sure you get the best possible experience from your interactions with the Russian culture and people. 1. Learn as much as you Can: \\
			\hline
			\small 3 & \small So, if you want to use the power of positive thinking to improve your life, start small and build up gradually. Believe in yourself and your abilities, and don't worry about criticism as it's an integral part of growth. \\
			\hline
			\small 4 & \small A New York City-based non-profit has been instrumental in fostering greater transparence in the city's supplementary food system. This has involved working with foodstores, restaurants, and community organizations to promote greater understanding and access to healthier foods. \\
			\hline
			\small 5 & \small The global COVID-19 pandemic has significantly influenced the worldwide popularity of online platforms and digital services. \\
			\hline
		\end{tabular}
	}
	\label{tab:sample_6x2_centered}
\end{table*}

\section{More experimental results}
\label{sec:more_exp_results}

\begin{figure}[h!]
	\vspace{-.1cm}
	\setlength{\abovecaptionskip}{-.1cm}
	\setlength{\belowcaptionskip}{-.3cm}
	\centering{
		\includegraphics[width=0.5\columnwidth]{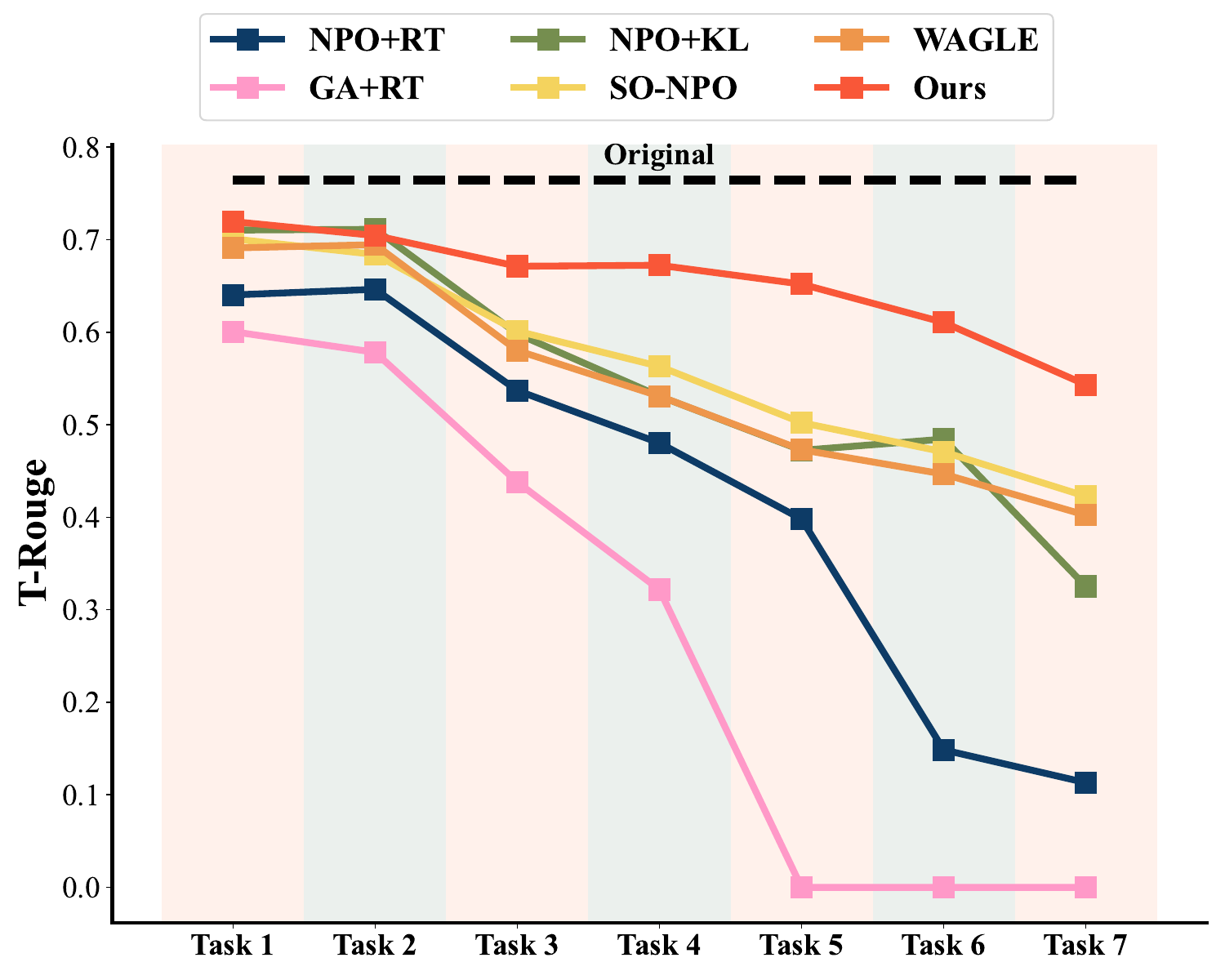}}
	\caption{T-Rouge across seven continual unlearning tasks on the mixed dataset of MUSE News and WHP. Different background color represents tasks from different datasets.}
	\vspace{-.3cm}
	\label{fig:mixed_dataset}
\end{figure}

\subsection{Mixed dataset.}
We argue that the higher the data relevance between unlearning tasks, the more severe the cascading degradation phenomenon. To evaluate the performance of our method in scenarios with low data relevance in continual unlearning, we conduct experiments mixing tasks from the MUSE News and WHP datasets.
In these experiments, the model is sequentially tasked to unlearn data from the two datasets, and the results are shown in Figure~\ref{fig:mixed_dataset}.
After completing the first MUSE News task, the model proceeds to unlearn the second WHP task, during which all methods exhibit a slight decline in model utility.
The decline is less severe compared to unlearning consecutive WHP tasks.
Similarly, our method maintains stable model utility when unlearning unrelated tasks sequentially.
When moving to the third task—another MUSE News task—baseline methods experienced significant performance degradation due to the cascading degradation effect initiated by the first task.
In contrast, our method demonstrates its robustness by effectively mitigating this phenomenon and maintaining model utility.

\subsection{Experiments on the TOFU dataset}
\begin{figure}[h!]
	\centering
		\begin{minipage}{0.7\columnwidth} %
			\makeatletter\def\@captype{table}
			\caption{Max F-Rouge across five unlearning tasks on TOFU dataset for Llama2 model.}
			\resizebox{\textwidth}{!}{
					\centering
					\begin{tabular}{c|cccccc}
							\toprule
							Methods &  NPO+RT & GA+RT & NPO+KL & SO-NPO & WAGLE & Ours\\
							\hline
							Max F-Rouge$\downarrow$ & 0.3398 & 0.3167 & 0.4538 & 0.3614 & 0.4621 & 0.3067 \\
							\bottomrule
                        \label{tab:tofu_llama2_maxF}
						\end{tabular}			
				}
		\end{minipage}
		\begin{minipage}{0.7\columnwidth} %
				\makeatletter\def\@captype{figure}
				\includegraphics[width=\columnwidth]{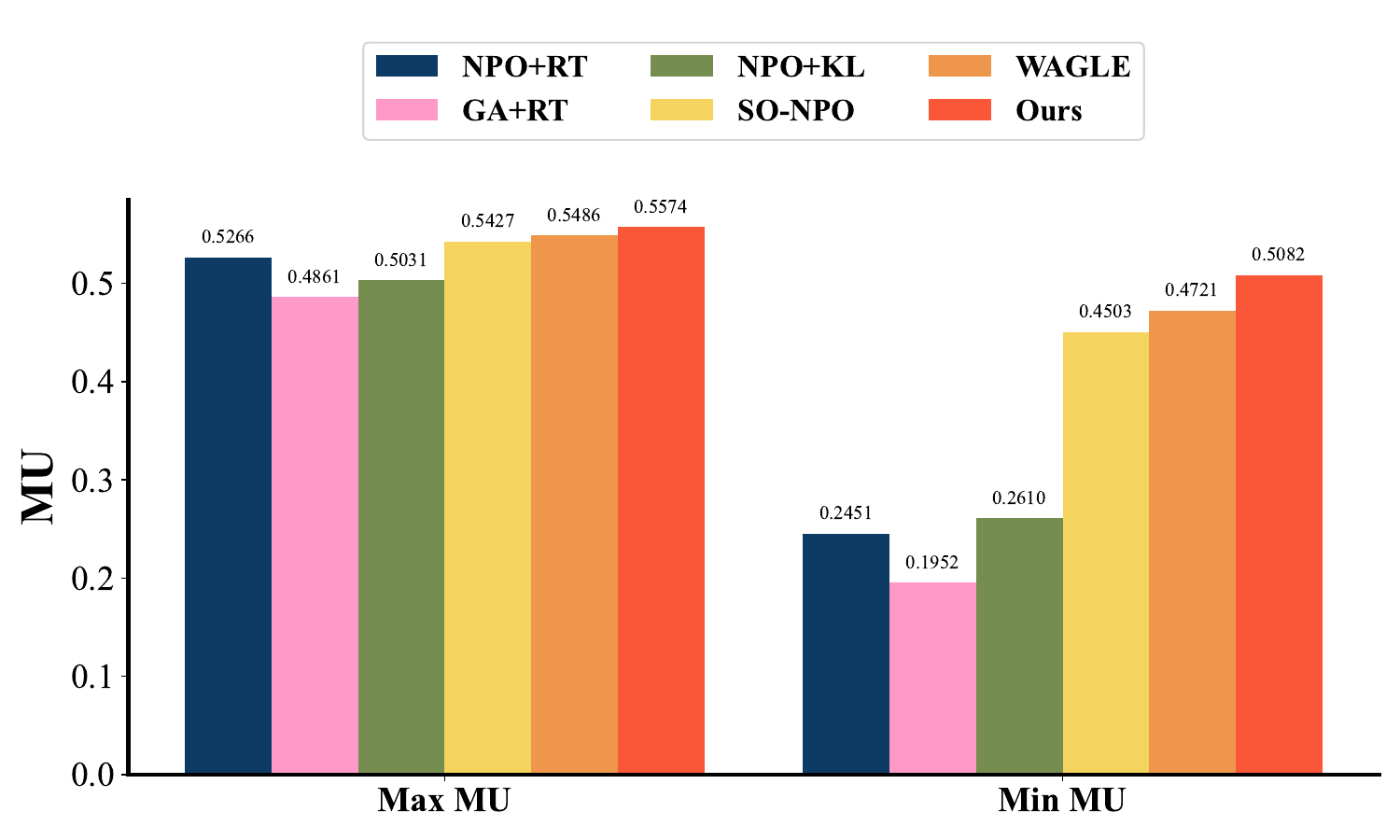} %
				\caption{Max/Min Model Utility (MU) across five unlearning tasks on TOFU dataset for Llama2 model.} %
                \label{fig:tofu_llama2}
			\end{minipage}
		
\end{figure}

We present maximum and minimum model utility across 5 unlearning tasks in Figure~\ref{fig:tofu_llama2}.
And the maximum F-Rouge is shown in Table~\ref{tab:tofu_llama2_maxF}, illustrating the performance of the least forgotten task after unlearning.
Our proposed method achieves the lowest F-Rouge, outperforming baseline methods in terms of unlearning performance.
Figure~\ref{fig:tofu_llama2} compares the ability of retaining model utility among different methods.
For Max MU, the proposed method attains the highest value (0.5574), followed by WAGLE (0.5486) and SO-NPO (0.5427). Similarly, for Min MU, the proposed method leads with a score of 0.5082, exceeding WAGLE (0.4721) and SO-NPO (0.4503), demonstrating superior retention of utility after unlearning all tasks. These results validate the proposed approach's ability to balance effective unlearning and high model utility.

\subsection{Experiments on the MUSE News dataset}

\begin{figure}[h!]
	\vspace{-.1cm}
	\setlength{\abovecaptionskip}{-.1cm}
	\setlength{\belowcaptionskip}{-.3cm}
	\centering{
		\includegraphics[width=0.9\columnwidth]{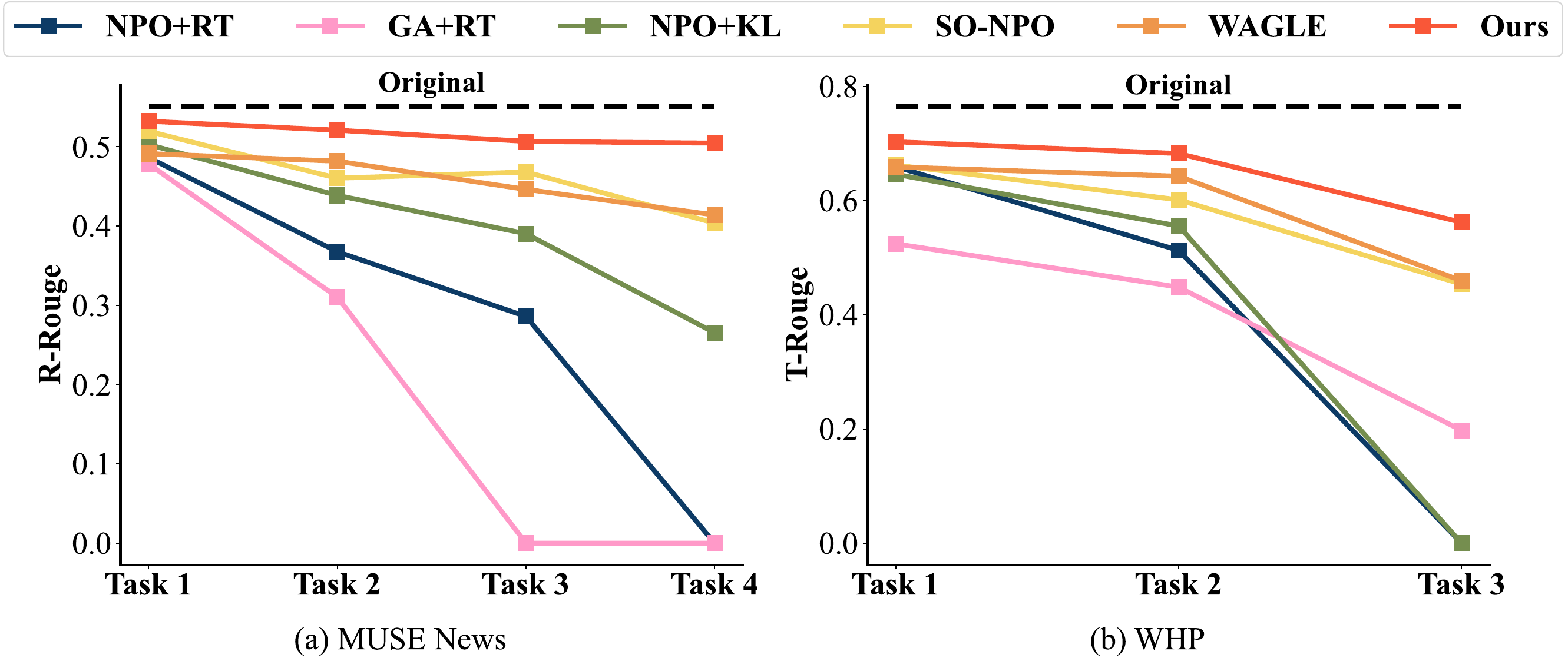}}
	\caption{(a) R-Rouge across four continual unlearning tasks on the MUSE News dataset. (b) T-Rouge across three continual unlearning tasks on the WHP dataset.}
	\label{fig:MUSE_News_WHP_rouge}
\end{figure}

\textbf{MUSE News.} Figure~\ref{fig:MUSE_News_WHP_rouge} (a) illustrates the utility and unlearning results on the MUSE News dataset, which exhibits more severe over-unlearning compared to the TOFU dataset.
Specifically, for baseline methods like NPO+RT and GA+RT, the utility drops to zero during the third or fourth unlearning task.
This can be attributed to the nature of the dataset, which requires verbatim unlearning of textual data rather than conceptual unlearning.
Verbatim unlearning poses a greater challenge and directly disrupts the generative capabilities of the models.
Moreover, the initial memorization of the model regarding the news content is relatively shallow; thus, high intensity and continuous unlearning can easily lead to excessive unlearning.
Our method, leveraging entropic soft labels, adaptively adjusts the unlearning intensity based on the prediction probabilities of the models for each sample. It ensures complete unlearning while dynamically modulating the unlearning intensity and selecting model parameters to avoid redundant unlearning.
As a result, our approach maintains model utility effectively, with negligible performance degradation after completing four unlearning tasks.

\subsection{Experiments on the WHP dataset}
Figure~\ref{fig:MUSE_News_WHP_rouge} (b) illustrates the T-Rouge performance of various methods across three continual unlearning tasks on the WHP dataset.
T-Rouge measures model utility on the TRAVIS dataset, which includes diverse genres, with higher values of T-Rouge indicating better retention of general knowledge.
The "Original" baseline, shown as the dashed line, represents the model's performance without unlearning.
Among the methods, our proposed approach demonstrates superior performance, maintaining the highest T-Rouge scores across all three tasks, indicating minimal degradation in utility.
While competing methods like WAGLE and SO-NPO exhibit better retention than NPO+RT and GA+RT, they still show significant drops in performance, especially as tasks progress. 
As shown in Table~\ref{tab:WHP_Trouge}, our method achieves comparable unlearning performance as other baselines.
The consistently higher T-Rouge scores of the proposed method reflect its ability to balance effective unlearning with the retention of general knowledge across diverse tasks, surpassing all baseline methods. This highlights its efficacy in handling challenging continual unlearning scenarios.

\subsection{Comparison between continual unlearning and one-time unlearning}
\begin{figure}[h!]
	\vspace{-.1cm}
	\setlength{\abovecaptionskip}{-.1cm}
	\setlength{\belowcaptionskip}{-.3cm}
	\centering{
		\includegraphics[width=0.5\columnwidth]{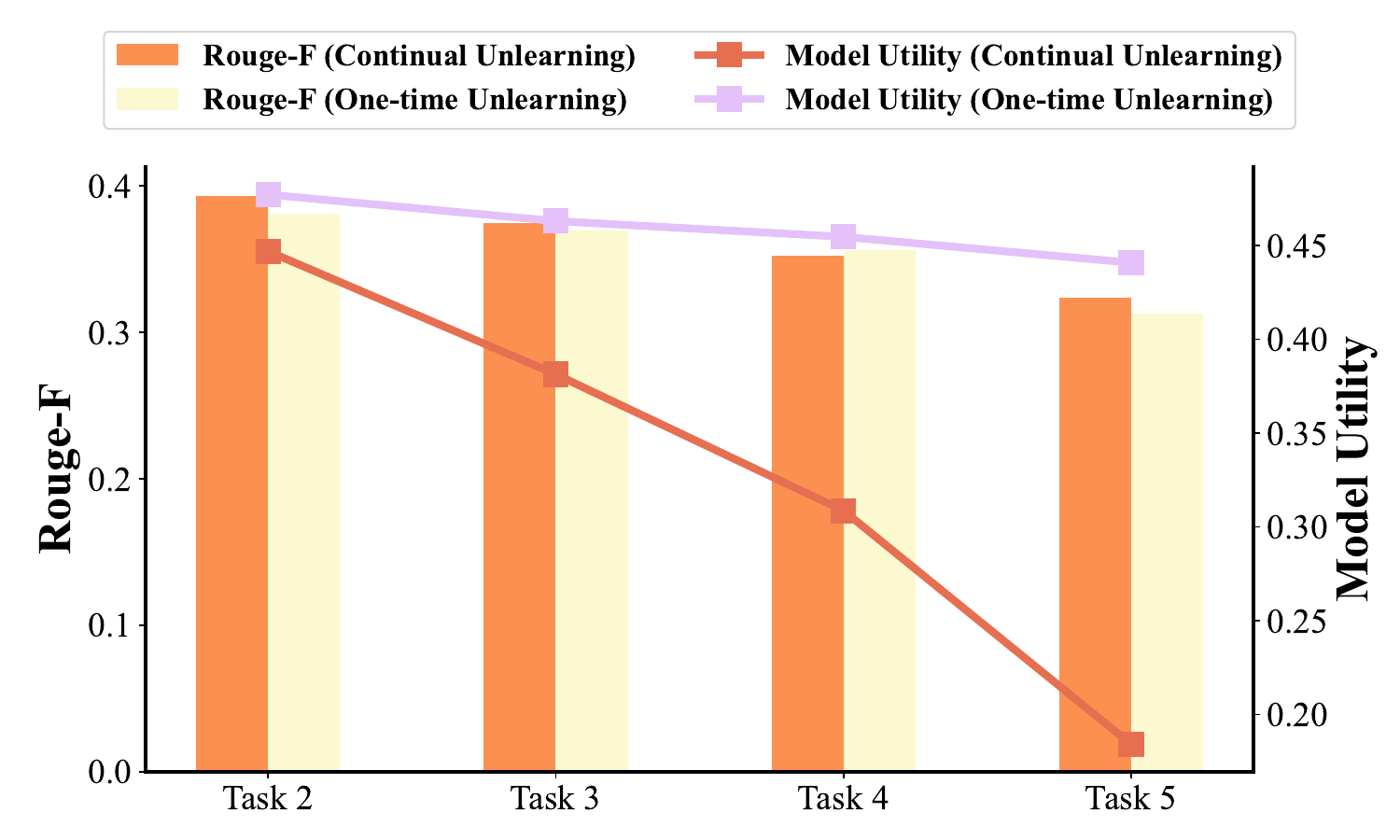}}
	\caption{Rouge-F and model utility under two unlearning scenarios.}
	\vspace{-.3cm}
	\label{fig:continual_onetime_unlearning_compare}
\end{figure}

To further validate the harmful effect of cascading degradation, we compare the model utility under continual unlearning and one-time unlearning scenarios in Figure~\ref{fig:continual_onetime_unlearning_compare}.
For one-time unlearning, all the data for current and previous unlearning tasks is jointly used for unlearning, which is only for experimental purposes and unrealistic in practical applications.

The Rouge-F in these two scenarios are aligned for a fair comparison of model utility.
As shown in the figure, continual unlearning results in a significantly more drastic decline in utility, while one-time unlearning results in a mild decrease in utility as the number of tasks increases.
This difference indicates the severe damage caused by cascading degradation in utility.

\end{document}